\documentclass{article}
\usepackage{textgreek}

\usepackage[preprint]{tmlr}

\usepackage[utf8]{inputenc}
\usepackage[T1]{fontenc}
\usepackage{hyperref}
\usepackage{url}
\usepackage{booktabs}
\usepackage{amsfonts}
\usepackage{nicefrac}
\usepackage{microtype}
\usepackage{xcolor}
\usepackage{graphicx}
\usepackage{amsmath}
\usepackage{amssymb}
\usepackage{multirow}
\usepackage{subcaption}
\usepackage{tabularx}
\usepackage{enumitem}
\usepackage{tikz}
\usetikzlibrary{arrows.meta,positioning,shapes.geometric,calc}
\usepackage{algorithm}
\usepackage{algpseudocode}
\usepackage{longtable}

\title{Long-Horizon Autonomous Architecture Research with a Language-Model Agent: A Behavioural Case Study}

\author{%
\begin{tabular}[t]{@{}l@{\hspace{1.5cm}}l@{}}
{\normalsize\bf Aon Safdar} & {\normalsize\bf Mohamed Saadeldin} \\
{\small\it University College Dublin, Ireland} & {\small\it University College Dublin, Ireland} \\
{\small\it aon.safdar@ucdconnect.ie} & {\small\it mohamed.saadeldin@ucd.ie}
\end{tabular}%
}

\def\month{08}
\def\year{2026}

\makeatletter
\def\@maketitle{\vbox{\hsize\textwidth \centering
{\LARGE\bf\sffamily \@title\par}\vskip \aftertitskip
{\@author\par}
\vskip 0.3in minus 0.1in}}
\makeatother

\begin{document}

\maketitle

\begingroup\footnotesize\itshape\noindent
This work has been submitted to the IEEE for possible publication. Copyright may be transferred without notice, after which this version may no longer be accessible.\par
\endgroup
\vspace{0.5em}

\begin{abstract}
We study what happens when a single general-purpose large language model is asked to act as the sole researcher on a long-horizon neural-architecture design problem. The agent is given a well-framed scientific question, an initial hypothesis and motivation, a compute allocation, and a small but carefully chosen set of research affordances (source and experiment management, experiment tracking, literature access, and a persistent working memory) and is then left to propose, implement, evaluate, and record its own experiments over an extended wall-clock horizon. We organise the programme into three phases separated by human-declared transitions, each phase expanding the agent's tool surface or the problem scale. Over approximately one hundred sequential single-variable experiments the agent drives a non-standard vision transformer from a weak baseline to a strong parameter-efficient design at small benchmark scales and to a usable but sub-SOTA design on ImageNet-1K, and generates a dense behavioural trace that we analyse here. We report four findings. (i)~The agent's productivity has a visible \emph{phase structure} (an early rapid-gain regime, a multi-dozen-hypothesis saturation wall, and a recovery regime) and the transition out of saturation is triggered by expanding the agent's action surface rather than by changing the underlying model. (ii)~A single early hypothesis accounts for roughly 80\% of the total CIFAR-10 accuracy gain, with the remainder long-tailed. (iii)~The agent's observed preference for greedy, incremental hypotheses is primarily \emph{workflow-induced}: a commit-or-discard evaluation rule is isomorphic to greedy hill-climbing, and the agent faithfully executes it; a residual component is attributable to risk aversion after bold failures and to anchoring on familiar literature. (iv)~The agent independently rediscovers several well-established results and, in the unfamiliar regime of pure channel attention, overturns a standard design choice. We conclude that, in this run, workflow design was at least as consequential as agent capability for long-horizon autonomous research, and we propose candidate modifications (diversified search, budgeted moonshot hypotheses, explicit forks, and regime-aware re-validation), offered as testable hypotheses for future autonomous-research systems.

\end{abstract}

\section{Introduction}
\label{sec:introduction}

Automated machine-learning research has traditionally been organised around a fixed search algorithm operating over an explicitly specified space: random or Bayesian hyperparameter search~\citep{bergstra2012random,snoek2012practical,li2017hyperband}, reinforcement-based or evolutionary neural architecture search~\citep{zoph2016nas,real2019regularized,liu2018darts,white2023nas}, or gradient-based weight-sharing methods. In each of these, the proposer, the evaluator, and the selector are discrete components and the search space is fixed in advance. A second wave of recent systems places a large language model in one of these roles, typically as the proposer or the code-writer, within an otherwise classical loop~\citep{romera2024funsearch,lu2024aiscientist,schmidgall2025agentlab,yang2024mle,yang2024sweagent,wang2024openhands}.

In this paper we take a further step: we give a single general-purpose language model \emph{all} of the research roles (proposer, implementer, executor, interpreter, and archivist) and study what it does when it is left to run for hundreds of GPU-hours and dozens of iterations on a substantive architectural problem. Our interest is not primarily in whether such an agent can substitute for a human researcher, but in what the agent's \emph{behaviour} looks like over a long horizon and what role the surrounding workflow plays in shaping it.

\paragraph{Setup.}
We equip the agent with four classes of research affordance:
source and experiment management for recording each hypothesis as a commit;
experiment tracking for monitoring training dynamics;
literature access (granted partway through the programme) for grounding hypotheses in prior work;
and a persistent structured working memory that records motivation, literature basis, implementation, budget, and outcome for each hypothesis. All job execution takes place on a shared-cluster workload manager; the agent submits one experiment at a time and ends its turn immediately after submission, with an external signal returning control when the job has finished. Human authorship is confined to the initial research brief, a small number of phase-level interventions, and occasional corrections of autonomy failures; all tactical decisions (which hypothesis to propose next, which code to change, which result to commit or discard) are made by the agent.

The research problem is the design of a channel-primary vision transformer operating under a pure channel-attention constraint (no spatial self-attention at any stage). This problem is substantive, constrained, and unfamiliar: standard ViT tricks target spatial attention, so the agent cannot fall back on well-known recipes. The architectural output of the loop is reported here only to the extent that it grounds the behavioural analysis; the object of study in this paper is the loop itself, not the resulting architecture.

\paragraph{Programme.}
We organise the run into three phases, separated by human-declared transitions at milestones. Each transition either expands the agent's tool surface (for example, granting literature access) or changes the problem scale (small classification benchmarks, then a larger one, then ImageNet-1K). Across these phases the agent proposes and evaluates roughly one hundred single-variable hypotheses, iterates the architecture from a weak baseline to a strong parameter-efficient design at small benchmark scales (and a usable but sub-SOTA design at ImageNet-1K scale), and produces for each hypothesis a time-stamped record with motivation, implementation, and outcome.

\paragraph{Contributions.}
Our contribution is the methodological study, not the architecture. Because our evidence is a single contiguous run of a single agent on a single architectural problem, we frame the findings below as \emph{scoped observations} from this case study rather than as general laws; we support each with a within-study quantitative decomposition (Sections~\ref{sec:quantitative}--\ref{sec:agent_behavior}), while flagging throughout that these are observational contrasts drawn from one trajectory, not randomised ablations across agents or problems (Section~\ref{sec:limitations}). Specifically:

\begin{enumerate}[topsep=2pt,itemsep=2pt,leftmargin=*]
    \item We describe a workflow that supports long-horizon, single-agent autonomous research on a real HPC allocation, including the persistence discipline, the structured memory, and the human-intervention record (Section~\ref{sec:system}).
    \item We quantitatively analyse the agent's hypothesis trajectory (Section~\ref{sec:quantitative}). We observe a three-regime productivity structure (early rapid gains, saturation, recovery), a strong concentration of benefit in the single earliest hypothesis, and a productivity phase transition that coincides with the addition of literature access (one of several simultaneous workflow changes; Section~\ref{sec:limitations}).
    \item We examine the agent's hypothesis-generation patterns and decompose its observed bias toward greedy, incremental proposals (Section~\ref{sec:agent_behavior}). We argue that most of the bias is induced by the evaluation rule of the workflow, with a residual component attributable to risk aversion after bold failures and to anchoring on the papers already familiar to the agent.
    \item We catalogue findings that the agent \emph{independently rediscovered} from a pool of well-known ML results, findings it \emph{did not discover} despite being explicitly named in the brief, and a finding from the standard design regime that the agent \emph{overturned} in the unfamiliar regime of pure channel attention (Section~\ref{sec:agent_behavior}).
    \item We propose candidate workflow modifications (diversified proposal, budgeted moonshot hypotheses, explicit architectural forks, and regime-aware re-validation) grounded in the observed failure modes and offered as testable hypotheses for future autonomous-research systems (Section~\ref{sec:lessons}).
\end{enumerate}

\paragraph{Why this matters.}
Recent LLM-agent benchmarks focus on short-horizon, task-completion settings where an agent is handed a single issue, patch, or dataset and is judged on outcome within hours~\citep{yang2024sweagent,jimenez2023swebench,wang2024openhands,yang2024mle}. The behavioural phenomena that dominate long-horizon research (saturation plateaus, anchoring, success-chasing, risk aversion after bold failures, carry-over of small-scale decisions into larger-scale regimes) are not visible at that time scale. We report them here as a first data point for designers of future autonomous-research systems: the interesting question is not whether a frontier LLM can \emph{in principle} do research, but which workflow discipline makes a flawed-but-tireless agent a productive research collaborator over weeks of wall-clock and hundreds of experiments.

\section{Related Work}
\label{sec:related_work}

\paragraph{Hyperparameter optimisation and neural architecture search.}
Classical AutoML automates either the hyperparameter search or the architecture search, using a fixed proposer (random, Bayesian, reinforcement, evolutionary, gradient-based weight-sharing) over an explicitly defined space~\citep{bergstra2012random,snoek2012practical,li2017hyperband,zoph2016nas,liu2018darts,real2019regularized,white2023nas}. Our setting dispenses with both the fixed proposer and the fixed space: the agent generates its own hypotheses in an open-ended fashion, drawing on an LLM's general knowledge of machine learning, and each hypothesis is realised as a code-level change and evaluated end-to-end. The phenomena we observe (saturation plateaus, anchoring on recent success, risk aversion after bold failures) are specific to open-ended proposal and are not well captured by the sample-efficiency metrics of classical NAS.

\paragraph{LLM agents for code and software engineering.}
A growing class of agent systems targets code-modification tasks, benchmarked primarily on SWE-bench-style issue-resolution traces~\citep{jimenez2023swebench,yang2024sweagent,wang2024openhands}. These agents typically build on general reasoning-and-acting and self-reflection scaffolds~\citep{yao2023react,shinn2023reflexion}, and operate on short horizons: a bug report arrives, a patch is produced, the trace ends. In our setting the agent must accumulate knowledge over dozens of hypotheses, track which regions of design space have been refuted, and adapt its proposal strategy as the space saturates. The behavioural failure modes we emphasise (saturation walls, greedy anchoring, incrementalism after a bold failure) are invisible at short-horizon benchmark time scales.

\paragraph{LLMs for scientific and ML-research discovery.}
Several recent systems couple LLMs to scientific or engineering objectives. FunSearch~\citep{romera2024funsearch} and the follow-on AlphaEvolve~\citep{novikov2025alphaevolve} use LLM-in-the-loop search for program and algorithm discovery; AI Scientist~\citep{lu2024aiscientist} produces end-to-end short research papers; MLE-bench~\citep{yang2024mle} and MLAgentBench~\citep{huang2024mlagentbench} evaluate agent behaviour on time-bounded ML engineering tasks; Agent Laboratory~\citep{schmidgall2025agentlab}, ResearchAgent~\citep{baek2024researchagent}, and the Virtual Lab~\citep{swanson2024virtuallab} explore multi-agent and idea-generation research workflows; Boiko et al.~\citep{boiko2023autonomous} report a long-horizon autonomous chemistry agent on a substantive scientific task. Our work differs in three respects. First, we focus on \emph{structural architecture research}, in which the agent writes and rewrites the model code itself, rather than on program synthesis, manuscript generation, hyperparameter optimisation, or time-bounded benchmark tasks. Second, our case study is substantially longer than most published LLM-research case studies, allowing long-horizon behavioural phenomena (saturation plateaus, anchoring, post-failure risk aversion, cross-scale anti-transfer) to appear and be quantified. Third, our primary contribution is \emph{descriptive and methodological}: we analyse what the agent does and which workflow choices shape it, rather than claim a general substitute for human researchers.

\paragraph{Human-AI research collaboration.}
A complementary line of work positions language models as research \emph{collaborators} rather than as autonomous researchers~\citep{messeri2024artificial,si2024llm,zheng2025survey}. Our framing is agnostic on this spectrum: most day-to-day decisions are taken by the agent, while human intervention is limited to strategic transitions and occasional reliability repairs, with every intervention recorded and reported.

\paragraph{Reproducibility.}
Because the behavioural trace is itself the primary object of study, we document the per-hypothesis record, the experiment logs, and the workflow in sufficient detail for other groups to audit the trace and replicate the analysis on different agents and workflows.

\section{The Autonomous Research Loop}
\label{sec:system}

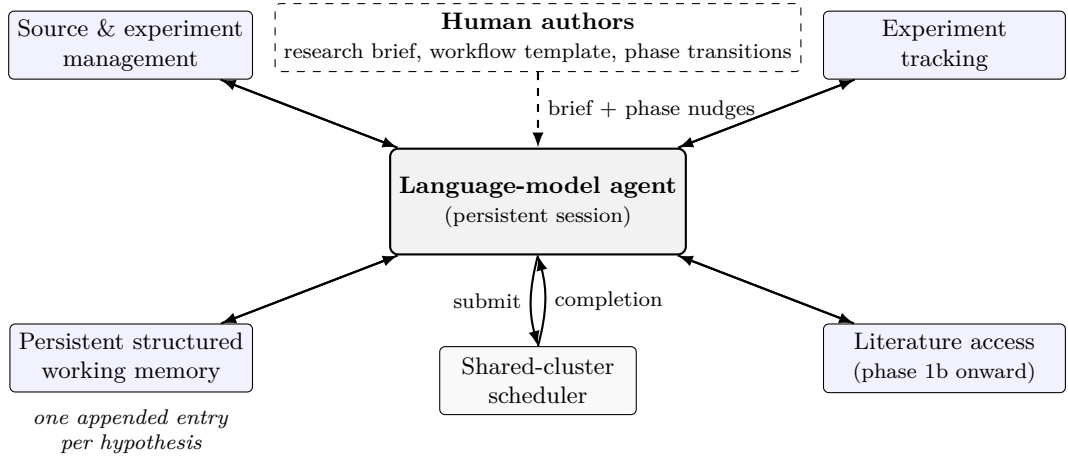
\begin{figure}[t]
\centering
\begin{tikzpicture}[
    node distance=6mm and 10mm,
    box/.style={draw, rectangle, rounded corners=2pt, minimum width=26mm,
                 minimum height=9mm, align=center, font=\small},
    agent/.style={draw, thick, rectangle, rounded corners=3pt, minimum width=36mm,
                    minimum height=14mm, align=center, fill=black!5, font=\small},
    aff/.style={draw, rectangle, rounded corners=2pt, minimum width=32mm,
                  minimum height=9mm, align=center, fill=blue!5, font=\small},
    human/.style={draw, dashed, rectangle, rounded corners=2pt, minimum width=40mm,
                    minimum height=9mm, align=center, font=\small},
    arrow/.style={-{Latex[length=2mm]}, thick},
    darrow/.style={-{Latex[length=2mm]}, thick, dashed},
]
% Central agent
\node[agent] (agent) {\textbf{Language-model agent}\\\footnotesize (persistent session)};

% Research affordances -- arranged around the agent
\node[aff, above left=9mm and 18mm of agent] (src) {Source \& experiment\\management};
\node[aff, above right=9mm and 18mm of agent] (track) {Experiment\\tracking};
\node[aff, below left=9mm and 18mm of agent] (mem) {Persistent structured\\working memory};
\node[aff, below right=9mm and 18mm of agent] (lit) {Literature access\\\footnotesize (phase 1b onward)};

% Compute / scheduler
\node[box, below=12mm of agent, fill=gray!5] (sched) {Shared-cluster\\scheduler};

% Human intervention (dashed box)
\node[human, above=10mm of agent] (human) {\textbf{Human authors}\\\footnotesize research brief, workflow template, phase transitions};

% Arrows: agent <-> affordances
\draw[arrow] (agent) -- (src);
\draw[arrow] (src) -- (agent);
\draw[arrow] (agent) -- (track);
\draw[arrow] (track) -- (agent);
\draw[arrow] (agent) -- (mem);
\draw[arrow] (mem) -- (agent);
\draw[arrow] (agent) -- (lit);
\draw[arrow] (lit) -- (agent);

% Agent -> scheduler (submit), scheduler -> agent (completion)
\draw[arrow] (agent.south) to[bend right=15] node[left, font=\footnotesize] {submit} (sched.north);
\draw[arrow] (sched.north) to[bend right=15] node[right, font=\footnotesize] {completion} (agent.south);

% Human -> agent
\draw[darrow] (human.south) -- node[right, font=\footnotesize] {brief + phase nudges} (agent.north);

% Per-hypothesis output back to memory (annotation)
\node[font=\footnotesize\itshape, below=1mm of mem, align=center, text width=35mm]
    {one appended entry\\per hypothesis};

\end{tikzpicture}
\caption{The autonomous research loop. A single persistent language-model agent is granted four research affordances (source/experiment management, experiment tracking, persistent structured working memory, and, from the second phase onward, literature access) and submits one experiment at a time to a shared-cluster scheduler. Human authorship is confined to the research brief, the workflow template, and a small number of phase-transition interventions; all tactical decisions (hypothesis selection, code changes, result interpretation, commit-or-discard) are taken by the agent.}
\label{fig:loop_diagram}
\end{figure}

\subsection{Design Principles}
\label{sec:system_principles}

We built the loop around four principles derived from the failure modes of earlier unstructured attempts; Figure~\ref{fig:loop_diagram} depicts the resulting loop.

\emph{Single agent, single hypothesis.} At any time the agent is executing exactly one experiment. We intentionally do not parallelise: parallel execution dilutes the causal signal from each change and makes failure diagnosis harder. The cost of this discipline is wall-clock; the benefit is an unambiguous per-hypothesis outcome that the agent can reason about.

\emph{Commit-or-discard as the evaluation rule.} After each experiment, the agent compares the result against the current champion. If improved, the change is preserved in source control and becomes the new baseline; otherwise the experimental change is reverted and the next hypothesis begins. We will return to this choice in Section~\ref{sec:behavior_bias}: it is operationally convenient and it structurally imposes greedy hill-climbing, which has observable downstream effects on the agent's hypothesis distribution.

\emph{Structured persistent memory.} Every hypothesis is recorded in a single research log in a fixed template that captures motivation, literature grounding, the exact change applied, the parameter and compute budget, and the result. The log is both the agent's primary cross-session memory and our primary unit of observation. We require the agent to read the log at the start of each session and to append to it before submitting each experiment.

\emph{End-of-turn after submission.} The agent is instructed to end its turn immediately after submitting a training job, rather than to poll the scheduler itself. Control is returned when the job leaves the queue. This discipline keeps the agent's token budget focused on scientific reasoning and avoids one class of autonomy failure; the concrete mechanism by which control is returned is an implementation detail described in Appendix~\ref{sec:appendix_impl}.

\begin{algorithm}[t]
\caption{The per-hypothesis autonomous research loop. One pass corresponds to one experiment.}
\label{alg:loop}
\begin{algorithmic}[1]
\State read research log $L$ and synthesis document $S$ (if present)
\State optionally query literature tools $T_\mathrm{lit}$ for prior art
\State propose a single-variable hypothesis $h$, with written motivation and literature grounding
\State edit code and/or configuration to realise $h$; verify parameter and compute budget
\State append a pre-run entry to $L$: motivation, literature, change, budget
\State submit one experiment to the cluster scheduler and end the turn
\Statex \quad\textit{(control returns to the agent on job completion)}
\State parse the training log for the primary metric $m$
\State append the post-run result to $L$: $m$, $\Delta$ vs current champion, short interpretation
\If{$m$ improves the current champion}
    \State commit the change to source control; the champion becomes the current state
\Else
    \State revert the change; the champion is unchanged
\EndIf
\State go to line 1
\end{algorithmic}
\end{algorithm}

\subsection{Research Affordances}
\label{sec:system_affordances}

The agent is given four classes of affordance. Each is intentionally minimal.

\emph{Source and experiment management.} A single version-control repository with a branching discipline: one branch per phase, one commit per accepted hypothesis. Rejected hypotheses are reverted from the working tree but preserved in the log.

\emph{Experiment tracking.} A standard experiment-tracking service records per-epoch metrics for every run. The agent is expected to reference training curves when interpreting ambiguous results.

\emph{Literature access.} From the second phase onward the agent can query preprint servers, code repositories with associated documentation, and a model-and-dataset hub, via a uniform tool protocol. Prior to this, only internal reasoning is available.

\emph{Compute.} A shared-cluster workload manager with a mix of accelerator generations. The agent submits jobs through this scheduler only, and is forbidden from running training on the interactive login host. Table~\ref{tab:compute} summarises the realised compute footprint of the run: approximately ten weeks of wall-clock and on the order of $2.4$\,k GPU-hours across roughly one hundred and fifty submitted jobs. The median substantive run is short ($\approx 2.5$\,h), with a long right tail dominated by the multi-day full-recipe ImageNet validation runs.

\begin{table}[h]
\centering
\caption{Compute footprint of the autonomous-research programme. Figures are rounded; jobs that exited within seconds of startup (e.g.\ misconfigured launches) contributed negligibly to total compute and are not counted as substantive runs. The shared cluster on which the run was executed provided a mix of accelerator generations; we report aggregate GPU-hours rather than per-accelerator usage.}
\label{tab:compute}
\small
\begin{tabular}{lr}
\toprule
Metric & Value \\
\midrule
Programme span (wall-clock)                & $\approx 10$ weeks \\
Training jobs submitted                    & $\approx 150$ \\
Substantive training runs ($\geq 5$ min)   & $\approx 130$ \\
Total GPU-hours                            & $\approx 2{,}400$ \\
Median wall-clock per substantive run      & $\approx 2.5$\,h \\
Longest single run                         & $\approx 6$ days \\
\bottomrule
\end{tabular}
\end{table}

\subsection{Human-Authored Interface}
\label{sec:system_interface}

Three human-authored documents bound the agent's behaviour. The first is the \emph{research brief}: the scientific question, the design constraints (for example, the pure channel-attention constraint in our case study), the compute and parameter budget, and a list of illustrative directions. The second is the \emph{workflow template}: the per-hypothesis loop described above, the expected structure of log entries, and explicit prohibitions (no scheduler polling, no login-node training, one job at a time). The third is a \emph{literature-grounding requirement}: every hypothesis must either cite a concrete prior result or explicitly declare that it is an internal extrapolation. We found each of these three documents to have a measurable effect on agent behaviour; the appendix details the specific wording that survived iteration.

\subsection{Human Intervention Record}
\label{sec:system_interventions}

We enumerate every human intervention across the programme. Interventions are bounded and strategic: they declare phase transitions, repair reliability failures, or expand the tool surface. They never make tactical decisions (hypothesis selection, hyperparameter choice, code writing, result interpretation), which remain the agent's responsibility.

\begin{table}[h]
\centering
\caption{Complete record of human interventions during the research programme. All interventions are strategic; all tactical decisions (hypothesis selection, configuration editing, code writing, and result interpretation) were made by the agent.}
\label{tab:interventions}
\small
\begin{tabularx}{\textwidth}{llX}
\toprule
When & Type & What changed \\
\midrule
Phase 1 start & Research brief issued & Scientific question, design constraints, parameter and compute budget, workflow template. \\
During Phase 1 & Reliability repair & Brief amended to instruct the agent to end its turn after submission rather than poll the scheduler itself. \\
Phase 1 / 1b boundary & Tool-surface expansion & Literature access enabled; brief amended to require literature grounding and to permit code-level changes. \\
Phase 1b / 2 boundary & Scale transition & Parameter budget raised; dataset difficulty increased; scale-transition nudge (``re-tune training recipe at the new scale''). \\
During Phase 2 & Research-direction nudge & Plateau flagged; brief amended to prioritise structural innovation over configuration tuning. \\
During Phase 2 & Memory discipline & Full research-log read required at session start; compact agent-authored synthesis document introduced. \\
Phase 2 / 3 boundary & Novelty discipline & Brief amended with a ``bold structural hypothesis'' requirement, a mandatory novelty self-audit per entry, and an illustrative menu of five example directions (funnelling, low-rank factorisation, sparse top-$k$, hierarchical local + global, linear / SSM channel mixers); we treat the menu as a documented direction-shaping intervention, since the agent's Phase 3 hypothesis set ended up including adaptations of all five. \\
Phase 3 start & Benchmark transition & Primary benchmark switched to ImageNet-1K, with a screening and a validation protocol. \\
End of Phase 3 & Programme closure & Programme stopped by the team after eighteen Phase-3 hypotheses, motivated by an observed saturation pattern around a low-rank attention champion and a desire to close out the behavioural analysis. No new research direction was prescribed. \\
\bottomrule
\end{tabularx}
\end{table}

\subsection{Reproducibility}
\label{sec:system_reproducibility}

We do not claim exact run-level numerical reproducibility, because LLM stochasticity will vary the specific hypothesis trajectory. We do predict that the \emph{qualitative} structure (early rapid gains, a multi-dozen-hypothesis saturation plateau, a recovery regime triggered by expanded tool access) will reappear on comparable problems, and we invite others to test this prediction.

\section{Case Study}
\label{sec:case_study}

\subsection{Research Problem}
\label{sec:case_study_problem}

The agent was given a channel-primary vision transformer in which channels play the role of tokens and spatial positions play the role of features. This inverts the usual arrangement in vision transformers~\citep{vaswani2017attention,dosovitskiy2021vit,touvron2021deit}, connecting the backbone instead to the channel-recalibration and cross-covariance lineage of channel attention~\citep{hu2018senet,elnouby2021xcit}. The central scientific question is whether a backbone that eliminates spatial self-attention entirely, relying only on channel attention with convolutional spatial mixing, can be made competitive with standard spatial-attention vision transformers, and with efficient convolutional baselines~\citep{tan2019efficientnet,liu2022convnext}, under a modest parameter budget. We chose this problem as the setting for the present study for three reasons. It is \emph{substantive}: a genuine architectural question with published prior art to orient against. It is \emph{constrained}: the agent cannot retreat to known ViT recipes, because the dominant published recipes target spatial attention. And it is \emph{unfamiliar}: the general-purpose LLM has seen spatial attention many times more than channel attention in its training data, so its intuitions are likely to be partly wrong, providing an opportunity to observe whether the agent notices, and how it responds.

\subsection{Phases}
\label{sec:case_study_phases}

The programme is organised into three phases, separated by human-declared transitions (Table~\ref{tab:phases}). Each transition either expanded the agent's tool surface (literature access in the second phase) or changed the problem scale (larger parameter budget and harder dataset, then ImageNet-1K). The agent did not choose these transitions; it operated autonomously within each phase.

\begin{table}[h]
\centering
\caption{The three research phases of the programme, distinguished by tool access and problem scale. The agent's observed behaviour differs visibly between phases (Section~\ref{sec:quantitative}).}
\label{tab:phases}
\small
\begin{tabular}{llcclc}
\toprule
Phase & Hypotheses & Dataset & Scale & Tool surface & Success rate \\
\midrule
1  & H1--H42       & CIFAR-10    & 5.8M  & config/code editing, tracking, source control & 36\% \\
1b & H43--H67      & CIFAR-10    & 5.8M  & + literature access (preprints, repositories, hub) & 38\% \\
2  & P2-H1--P2-H32 & CIFAR-100   & 22M   & full toolkit, larger budget              & 32\% \\
3  & H3-1--H3-18   & ImageNet-1K & 22M   & + multi-GPU distributed configurations  & 12\% \\
\bottomrule
\end{tabular}
\end{table}

\subsection{Headline Outcomes}
\label{sec:case_study_numbers}

Table~\ref{tab:case_study_results} reports the end-of-phase top-1 validation accuracy of the agent's architecture. The complete per-hypothesis trajectory is visualised in Figure~\ref{fig:trajectory} and tabulated in Appendix~\ref{sec:appendix_full_log}. We report architectural numbers only to the extent they ground the behavioural analysis; the object of study is the loop, not the final architecture, and the numbers are not intended as a stand-alone architectural claim.

\begin{table}[h]
\centering
\caption{End-of-phase top-1 validation accuracy for the agent-produced architecture. All CIFAR-10 / CIFAR-100 numbers are for the agent's directly iterated protocol; ImageNet numbers are Phase-3 validation. The Phase 3 champion row records the agent's best 100-epoch result; its 300-epoch validation was not run because the team elected to stop the programme at the saturation point rather than burn an additional week of H100 time on a $+0.26$\,pp delta.}
\label{tab:case_study_results}
\small
\begin{tabular}{lllrcc}
\toprule
Phase & Dataset & Identifier & Params (M) & Top-1 (100ep) & Top-1 (300ep) \\
\midrule
1 baseline  & CIFAR-10    & Baseline                          & 1.72  & 69.67\% & --- \\
1 champion  & CIFAR-10    & H30                               & 5.99  & 95.10\% & --- \\
1b champion & CIFAR-10    & H64                               & 5.81  & 96.59\% & 97.25\% \\
2 champion  & CIFAR-100   & P2-H26                            & 22.15 & 83.37\% & 85.07\% \\
3 baseline  & ImageNet-1K & IN-B100 (P2-H26 arch)             & 22.15 & 77.65\% & 79.00\% \\
3 champion  & ImageNet-1K & H3-12 (Linformer rank-128 stage-3) & 26.57 & 77.91\% & --- \\
\bottomrule
\end{tabular}
\end{table}

The overall CIFAR-10 improvement of roughly twenty-seven percentage points is dominated by a single early hypothesis (Section~\ref{sec:quantitative_attribution}). All subsequent gains are refinements. We regard this concentration of benefit as itself a data point about LLM-driven open-ended research: the largest gain is accessible almost immediately, and the remainder is long-tailed.

Phase 3 produces a qualitatively similar but compressed pattern at the ImageNet-1K scale. Eighteen hypotheses yield only two accepted improvements; the larger of them, an in-trunk low-rank channel-attention block adapted from the published Linformer mechanism~\citep{wang2020linformer}, carries the architecture from a $77.65\%$ baseline to a $77.91\%$ champion at $100$ epochs. The six hypotheses that follow (five narrow modifications of this champion and one attempt at a structurally different operator) all fail. The team closed Phase 3 at this point rather than continue iterating; the team also did not run the $300$-epoch validation of the Phase 3 champion, on the judgment that the resulting paper number would not change the behavioural story (Section~\ref{sec:agent_behavior}). The $300$-epoch validation of the Phase 3 \emph{baseline}, by contrast, was completed and is reported in Table~\ref{tab:case_study_results} as a calibration point: full training adds roughly $+1.35$\,pp on the same architecture, comparable to the $+1.70$\,pp the team observed for the Phase 2 champion under the same protocol.

\begin{figure}[t]
\centering
\includegraphics[width=\linewidth]{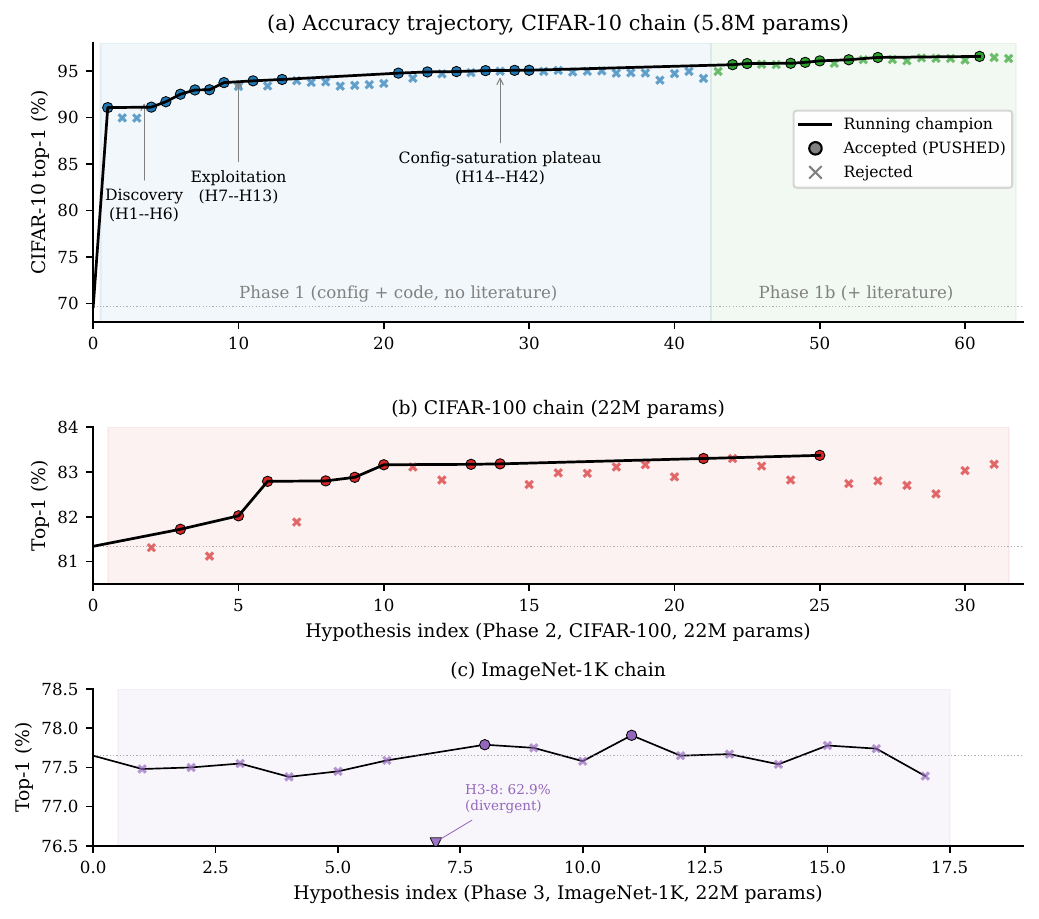}
\caption{Accuracy trajectory across the programme, broken out by dataset chain. (a) CIFAR-10 at 5.8M parameters, spanning Phase 1 and Phase 1b; (b) CIFAR-100 at 22M parameters, Phase 2; (c) ImageNet-1K at 22M parameters, Phase 3 (closed at H3-18). Circles mark accepted hypotheses (source-controlled as new champion); crosses mark rejected hypotheses (experimental change reverted). The solid black line is the running champion. Regime annotations on (a) highlight the early rapid-gain sub-regime, the short exploitation sub-regime, and the long configuration-saturation plateau that motivated Phase 1b. Panel (c) shows a $+0.26$\,pp aggregate gain over eighteen hypotheses with only two accepted, dominated by a single late acceptance (H3-12) and surrounded by a long sequence of stage-3 modifications that did not transfer.}
\label{fig:trajectory}
\end{figure}

\subsection{Outputs of the Loop}
\label{sec:case_study_output}

In addition to the accuracy improvement, the agent produced the following by-products of the loop: a complete architectural specification (embedding widths, block counts, normalisation placement, positional encoding, auxiliary supervision, residual shortcuts, and training recipe); a time-stamped research log of approximately one hundred hypotheses with motivation, literature grounding, change description, budget, and outcome; per-hypothesis source-control history that permits any intermediate design to be reconstructed; complete training telemetry for every experiment; and an agent-authored empirical-laws synthesis summarising what the programme had learned about channel-attention design. We treat these as artefacts of the loop; this paper's focus is on the loop that produced them and on the behavioural trace they reveal.

\section{Quantitative Analysis of the Hypothesis Trajectory}
\label{sec:quantitative}

We treat the per-hypothesis research log as a data set and analyse it along three axes: how success rate evolves across the programme, how total accuracy gain is distributed across individual hypotheses, and how productivity changes at tool-access and scale boundaries.

\subsection{Productivity Regimes}
\label{sec:quantitative_regimes}

The first phase of the programme decomposes cleanly into three sub-regimes of sharply different productivity (Table~\ref{tab:regimes}). An early \emph{discovery regime} of six hypotheses carries the agent from the initial baseline to within a few points of the first phase's ceiling, with very large per-hit gains. A short \emph{exploitation regime} of seven hypotheses, in which the agent tunes well-understood training hyperparameters and reorganises the hierarchy into four stages, accepts the majority of its proposals but with small per-hit gains; the only rejections are two over-aggressive learning-rate increases past the agent's own previously accepted setting. A long \emph{saturation regime} of twenty-nine hypotheses follows in which the agent continues to propose plausible configuration-level variations but few help, and the few that do help contribute only fractional points.

\begin{table}[h]
\centering
\caption{Phase-1 sub-regimes of productivity. Three patterns emerge: rapid architectural discovery (few hypotheses, very large per-hit gains), training-recipe exploitation (moderate success rate, small per-hit gains), and config-space saturation (many hypotheses, low success rate, near-zero per-hit gains). The third regime is the ``wall'' that literature-tool access (Phase 1b) was needed to break.}
\label{tab:regimes}
\small
\begin{tabular}{lllcc}
\toprule
Regime & Hypotheses & Description & Success rate & Avg gain / success \\
\midrule
Discovery    & H1--H6   & Architecture fundamentals (MLP, DW, depth) & 67\% (4/6)   & $+5.71$~pp \\
Exploitation & H7--H13  & Training recipe + 4-stage topology          & 71\% (5/7)   & $+0.32$~pp \\
Saturation   & H14--H42 & Topology + config tuning, config-space exhaustion & 21\% (6/29) & $+0.17$~pp \\
\bottomrule
\end{tabular}
\end{table}

\begin{figure}[t]
\centering
\includegraphics[width=\linewidth]{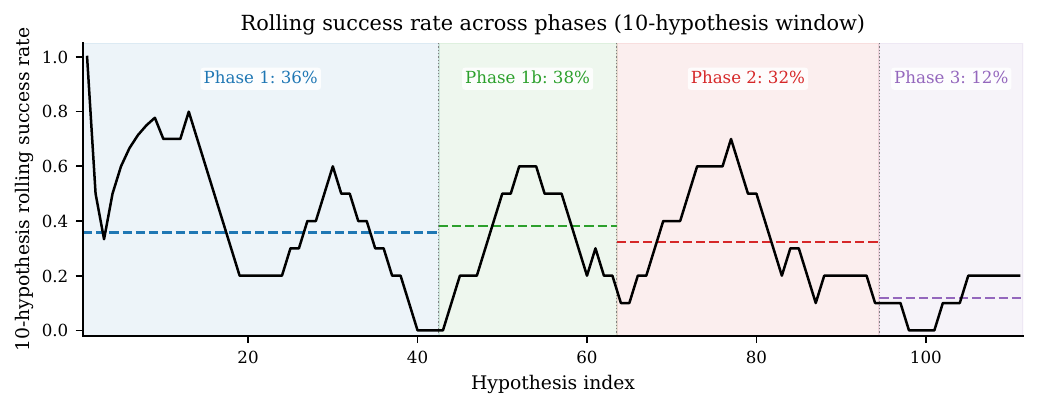}
\caption{Rolling ten-hypothesis success rate across the programme. Dashed horizontal lines give the per-phase mean success rate. Productivity drops sharply after the exploitation regime in Phase 1, does not recover until the Phase 1b transition that expanded the action surface, rises again after the Phase 2 scale transition, and collapses to $12\%$ in Phase 3 as the agent saturates around a Linformer-family champion (Section~\ref{sec:behavior_vignette}).}
\label{fig:success_rate}
\end{figure}

The saturation regime is the central behavioural phenomenon of long-horizon autonomous research (Figure~\ref{fig:success_rate}): LLM agents mine finite design spaces efficiently but lack a principled signal to declare a space exhausted and to escalate the kind of hypothesis being proposed. The agent does not run out of ideas; it runs out of \emph{useful} ideas in the current modality, and does not detect the shift on its own. In our programme, exit from saturation required a human-declared transition that expanded the action surface.

\subsection{Innovation Attribution}
\label{sec:quantitative_attribution}

\begin{figure}[t]
\centering
\includegraphics[width=\linewidth]{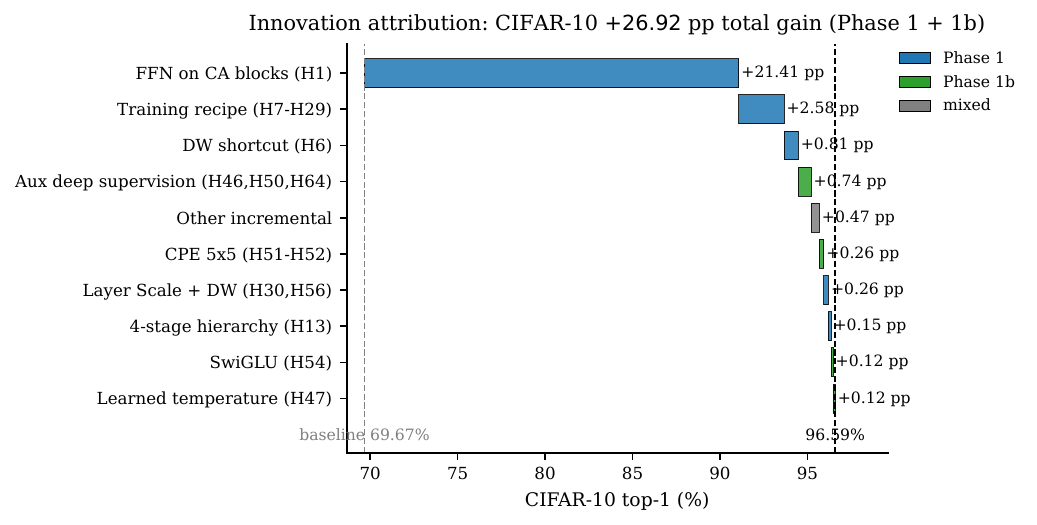}
\caption{Innovation attribution for the CIFAR-10 chain. The total $+26.92$\,pp improvement over the baseline decomposes into a single dominant contribution (the feedforward sublayer added by the first accepted hypothesis) and a long tail of small refinements, split between Phase-1 training-recipe configuration tweaks (the larger share by accumulated points) and Phase-1b literature-derived code changes (the more diverse mechanism-level innovations).}
\label{fig:attribution}
\end{figure}

Attributing the overall CIFAR-10 improvement of twenty-seven percentage points to individual innovations yields a strongly long-tailed distribution (Table~\ref{tab:attribution}; Figure~\ref{fig:attribution}). The single largest gain is produced by the first accepted hypothesis (the introduction of a feedforward sublayer on every channel-attention block) and accounts for roughly four-fifths of the total gain. Every subsequent accepted hypothesis is a refinement with a comparatively modest delta.

\begin{table}[htbp]
\centering
\caption{Innovation attribution on the CIFAR-10 $+26.92$~pp total gain. The single largest gain (H1, adding an FFN to channel-attention blocks) accounts for 79.5\% of the total. Long-tail gains are dominated by literature-guided Phase-1b code changes.}
\label{tab:attribution}
\small
\begin{tabular}{lrrl}
\toprule
Innovation & Gain (pp) & \% of total & Phase \\
\midrule
FFN (MLP) in CA blocks            & $+21.41$ & 79.5\% & Phase 1 (H1) \\
Training recipe (LR, warmup, aug) &  $+2.58$ & 9.6\%  & Phase 1 (H7--H29) \\
DW shortcut backbone              &  $+0.81$ & 3.0\%  & Phase 1 (H6) \\
Auxiliary deep supervision        &  $+0.74$ & 2.7\%  & Phase 1b (H46, H50, H64) \\
Other incremental                 &  $+0.47$ & 1.8\%  & various \\
CPE (5$\times$5 DW conv)          &  $+0.26$ & 1.0\%  & Phase 1b (H51--H52) \\
Layer Scale + DW fix              &  $+0.26$ & 1.0\%  & Phase 1/1b (H30, H56) \\
4-stage architecture              &  $+0.15$ & 0.6\%  & Phase 1 (H13) \\
SwiGLU activation                 &  $+0.12$ & 0.4\%  & Phase 1b (H54) \\
Learned attention temperature     &  $+0.12$ & 0.4\%  & Phase 1b (H47) \\
\bottomrule
\end{tabular}
\end{table}

Three observations follow. First, the largest available gain was accessible almost immediately: the agent did not need to explore to find it. Second, the long tail of subsequent gains is split: roughly two-thirds comes from training-recipe configuration tweaks proposed in Phase 1 (without literature access), and roughly one quarter comes from literature-derived code changes proposed in Phase 1b. The configuration tweaks are the larger share by accumulated points, but the literature-derived code changes are the more diverse and mechanism-level innovations. Third, the agent did not invent these ideas; it \emph{selected} them from the published literature and adapted them to the non-standard channel-attention regime. This selection-and-adaptation mode is what we would expect from a general-purpose LLM: its strength is broad but shallow coverage of the field, and it functions more as a well-read research assistant than as a source of novel mechanisms.

\subsection{Tool Access as a Phase Transition}
\label{sec:quantitative_tools}

The sharpest quantitative signal in the programme is the difference in productivity between the first and second phases (Table~\ref{tab:phase_transition}). Adding literature access did not raise the \emph{magnitude} of the best per-hypothesis gain (the largest gains had already occurred in the first phase) but it changed the \emph{kind} of hypothesis being proposed. The fraction of hypotheses that modified code rather than only configuration rose from a few percent to roughly three-quarters. Without this transition, the run would almost certainly have terminated at the first-phase ceiling.

\begin{table}[h]
\centering
\caption{Phase-1 $\to$ Phase-1b tool-access phase transition. Literature-tool access did not increase per-hypothesis gain magnitude, but it changed the \emph{kind} of hypothesis the agent proposed, and enabled the loop to break the config-saturation wall.}
\label{tab:phase_transition}
\small
\begin{tabular}{lrr}
\toprule
Metric & Phase 1 (no lit tools) & Phase 1b (with lit tools) \\
\midrule
Code changes (as fraction of hypotheses) & 2 / 42 = 5\%   & 16 / 21 = 76\% \\
Novel mechanisms introduced              & 2              & 8 \\
Distinct papers cited in proposals       & 0              & 24+ \\
Avg gain per successful hypothesis       & $+1.70$~pp     & $+0.19$~pp \\
Avg gain per successful hyp., excl. H1   & $+0.29$~pp     & $+0.19$~pp \\
\bottomrule
\end{tabular}
\end{table}

\begin{figure}[t]
\centering
\includegraphics[width=0.62\linewidth]{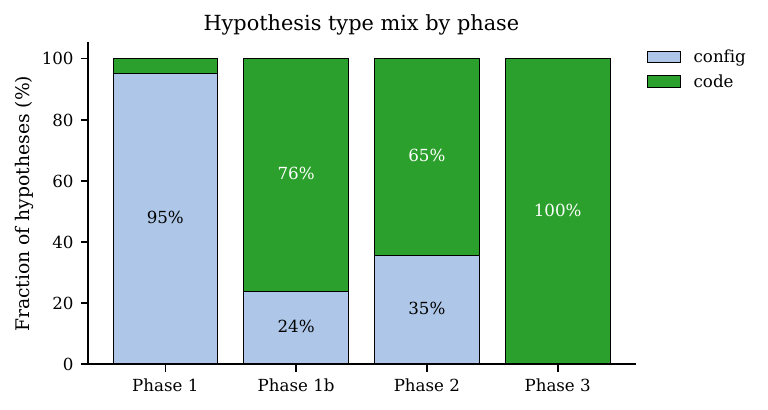}
\caption{Fraction of hypotheses whose only substantive change is a configuration toggle (light) vs a code-level architectural change (dark), per phase. The Phase 1b transition, which granted literature access and authorised code editing, shifts the agent's action from predominantly configuration to predominantly code.}
\label{fig:hypothesis_mix}
\end{figure}

We read this as evidence that the action surface made available to the agent is the single most important lever on long-horizon research productivity. The agent's underlying reasoning capability did not change across the boundary; what changed is what the agent was permitted, and equipped, to do.

\subsection{Cross-Scale Anti-Findings}
\label{sec:quantitative_cross_scale}

\begin{figure}[t]
\centering
\includegraphics[width=0.72\linewidth]{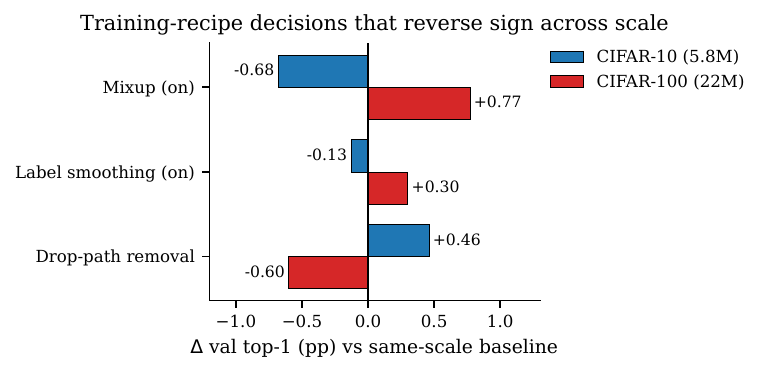}
\caption{Training-recipe decisions that reverse sign across scale. Positive bars indicate that turning the setting on helps at that scale; negative bars indicate that turning it off helps. Decisions that were beneficial on CIFAR-10 at 5.8M parameters become harmful on CIFAR-100 at 22M parameters, and vice versa.}
\label{fig:cross_scale}
\end{figure}

A subset of first-phase findings did not transfer to later phases, and in several cases reversed outright (Table~\ref{tab:cross_scale_reversals}; Figure~\ref{fig:cross_scale}). Settings that had been carefully tuned at the small benchmark were silently carried into the larger one, and required several hypotheses of re-tuning to recover. The agent did not automatically flag scale transitions; re-validation was only triggered by an explicit human nudge at the phase boundary. We include this as a cautionary finding for future autonomous-research systems: hyperparameter decisions validated at one scale should be treated as \emph{locally valid} and re-checked when the regime changes. The simplest mitigation is to annotate each accepted finding with the regime in which it was validated, and to require re-validation of the top findings whenever that regime changes.

\begin{table}[h]
\centering
\caption{Training-recipe decisions that reversed sign between CIFAR-10 (5.8M) and CIFAR-100 (22M). The agent had to re-learn each of these at the new scale. Any autonomous loop should track regime metadata and schedule automatic re-validation on scale transitions.}
\label{tab:cross_scale_reversals}
\small
\begin{tabular}{llll}
\toprule
Decision & CIFAR-10 (Phase 1) & CIFAR-100 (Phase 2) & Interpretation \\
\midrule
Mixup          & hurts (H21: $+0.68$~pp off) & helps (P2-H7: $+0.77$~pp on)   & virtual samples matter with 100 classes \\
Label smoothing & hurts (H23: $+0.13$~pp off) & helps (P2-H6: $+0.30$~pp on) & soft targets for fine-grained classes \\
Drop-path removal & helps (H7: $+0.46$~pp) & hurts (P2-H5: $-0.60$~pp)        & regularisation regime flips by capacity \\
\bottomrule
\end{tabular}
\end{table}

\section{Agent Behaviour}
\label{sec:agent_behavior}

The quantitative analysis tells us \emph{what} the agent produced. In this section we examine \emph{how} it proposed, reasoned, and stumbled.

\subsection{Hypothesis-Generation Patterns}
\label{sec:behavior_generation}

Six recurring patterns appear in the per-hypothesis log across the entire programme.

\emph{Systematic sweeping.} When the agent identifies a continuous dimension along which recent evidence suggests a gradient (learning rate, kernel size, depth balance) it sweeps the dimension one step at a time, typically overshoots, and then retreats. This behaviour is efficient for smooth objectives and is responsible for several of the long-tail gains in the second phase of the programme.

\emph{Ablation after success.} Immediately following an accepted hypothesis, the agent almost always proposes an ablation isolating the contribution of the new component. This is good scientific hygiene and we observe it consistently without having to prompt it.

\emph{Greedy anchoring on the current champion.} Every hypothesis in the second and third phases is written as an extension of the current champion. The agent never re-rooted its hypothesis tree on a discarded earlier branch, even during extended plateaus where doing so could have plausibly unlocked progress.

\emph{Context-sensitive retrial.} Several mechanisms were tried, rejected in one context, and re-tried in another: multi-head channel attention was tested twice at different depths; strong augmentation strategies were re-tested after a scale transition, succeeding in the new regime after failing in the old. This pattern suggests the agent is not simply memoising past failures but partially conditioning on regime.

\emph{Literature-grounded adaptation.} Once literature access was available, most hypotheses cited a specific paper and described the adaptation from the original setting to the present constrained setting. Quality varied: some adaptations succeeded cleanly, while others failed in informative ways (for example, a differential-attention variant that relies on competing spatial heads was tried with competing channel heads and performed worse than the baseline, an illustrative failure that we return to below).

\emph{Risk aversion after a bold failure.} After any prominently bold hypothesis that failed, the agent consistently retreated to incremental proposals for several consecutive turns, even when the workflow explicitly permitted another bold attempt. The agent did not resume bold proposals until externally prompted.

The last two patterns (greedy anchoring and post-failure risk aversion) together explain most of the agent's long-horizon drift toward incrementalism. They are the behavioural failure modes we most want to mitigate in future loops.

\subsection{Vignette: Saturation around a Phase 3 Champion}
\label{sec:behavior_vignette}

A clean miniature of the patterns above appears in the final ten hypotheses of Phase 3. The agent first accepts a low-rank channel-attention block adapted from the published Linformer mechanism~\citep{wang2020linformer}, placed in the deepest stage of the trunk (the Phase 3 champion; $+0.26$\,pp over the baseline at $100$ epochs). The next five hypotheses are all narrow variations on this same low-rank block: a wider rank schedule across stages, a per-block rank schedule, sparse selection over the learned basis, a parallel full-rank branch gated to zero at initialisation, and a sparsemax replacement for the softmax inside the projection. All five are rejected. The sixth hypothesis is the first one that proposes a structurally different operator (a Synthesizer-style dense attention map that bypasses key-query inner products entirely), and it crashes at initialisation on a distributed-data-parallel unused-parameter check; the agent diagnoses the unused projection itself, applies a one-line fix, and resubmits, without re-evaluating whether continuing to iterate in the same architectural neighbourhood is the right move at this point in the programme. The agent's per-hypothesis text in this stretch repeatedly flags the saturation explicitly (the cards include lines such as ``pivot to a fundamentally different operator family'') and then proposes another modification within the same neighbourhood on the very next turn. The instinct to escalate is present in the agent's verbal reasoning; the action consistently follows the gradient instead.

We read this stretch as a clean instance of the greedy-anchoring and risk-aversion patterns operating jointly: the loop has no mechanism to recognise the saturation other than as an after-the-fact pattern in the log, and the agent's own self-instructions to escalate do not override the local incentive to make a small, evaluable change. The team closed Phase 3 at this point.

\subsection{Independent Rediscoveries}
\label{sec:behavior_rediscovery}

Several well-established machine-learning findings emerged independently from the agent's experiments, without the agent having read the corresponding prior work in advance of the hypothesis (Table~\ref{tab:rediscovery}). The most instructive is the discovery that a \emph{single-head} channel attention outperforms multi-head in this regime. The standard intuition from spatial attention is the opposite: more heads are generally better. The agent tested multi-head channel attention twice, both times found it worse, and produced a clean post-hoc explanation grounded in the specific geometry of channel attention: splitting the spatial-feature dimension across heads weakens each head's ability to compute channel relationships. This is a case in which the agent resisted a transferred intuition from the standard regime and followed the experimental evidence instead.

\begin{table}[h]
\centering
\caption{Findings the agent converged on experimentally that are established or partially established in the ViT literature. The agent derived each of these from experimental evidence; the rightmost column lists the corresponding published result.}
\label{tab:rediscovery}
\small
\begin{tabularx}{\textwidth}{lXX}
\toprule
Finding & Agent's derivation & Established reference \\
\midrule
Single-head beats multi-head for channel tokens & H3 ($-1.13$~pp), H58 ($-0.36$~pp), confirmed twice & XCiT~\cite{elnouby2021xcit} uses single-head cross-covariance for the same reason \\
5$\times$5 is the Q/K kernel sweet spot at 7$\times$7 feature maps & H52 (5$\times$5 $>$ 3$\times$3), H53 (7$\times$7 worse) & Kernel-size / receptive-field correspondence; ConvNeXt-style~\cite{liu2022convnext} \\
Deeper $>$ wider at fixed parameter count & H26, H60, H15, P2-H2, all consistent & EfficientNet compound scaling~\cite{tan2019efficientnet} \\
Drop-path hurts underfitting models & H7 ($+0.46$~pp from removal) & Standard practice, but derived from first principles \\
RMSNorm $\neq$ LayerNorm for channel tokens & H57 ($-0.23$~pp) & Non-obvious; most work assumes equivalence \\
FFN is load-bearing even in CA-only models & H39 ($-1.07$~pp from removing FFN) & Universal Transformer conclusion, re-derived for channel attention \\
\bottomrule
\end{tabularx}
\end{table}

\subsection{Directions the Agent Did Not Pursue}
\label{sec:behavior_gaps}

Several directions that were explicitly enumerated in the research brief were never proposed by the agent under its own initiative: ensemble and multi-seed sensitivity analyses, curriculum and progressive training, state-space and linear-attention channel mixers, top-$k$ sparse channel attention, and knowledge distillation from a larger teacher. The agent read the brief containing these directions but consistently chose to iterate on dimensions where it had had recent success. We interpret this pattern as \emph{success-chasing}: the agent preferentially proposes hypotheses in the neighbourhood of its most recent accepted result. Success-chasing is operationally useful when the gradient is locally informative, but it is a strong source of drift on long horizons, and several of the gaps above were only closed after explicit human prompting.

\subsection{Instruction-Induced versus Inherent Bias}
\label{sec:behavior_bias}

A central question for any behavioural study of an LLM agent is the following. When the agent makes conservative, incremental choices, is it following a tendency of its own, or is it faithfully executing the workflow we imposed? We offer a partial decomposition.

\paragraph{The workflow is structurally a greedy hill-climb.}
The commit-or-discard evaluation rule, the single mutable working copy, the one-job-at-a-time queueing, and the ``build on the current champion'' anchoring template are jointly isomorphic to greedy local search: evaluate one neighbour, accept if better, revert otherwise. This workflow structurally forbids maintaining a population of diverse architectures, backtracking to an earlier branch point, or combining two individually failed ideas that might jointly succeed. Any agent executing it will display greedy behaviour regardless of its underlying preferences.

\paragraph{Evidence the bias is primarily workflow-induced.}
Three observations support this. First, at the literature-access transition the fraction of hypotheses modifying code jumped from a few percent to roughly three-quarters, without any change to the underlying agent: the action surface changed, the behaviour followed. Second, the agent \emph{did} propose backtracking when the workflow implicitly allowed it, but only as a contingent plan (``retry from an earlier branch if the forward path stalls''), suggesting the instinct exists but is not natively supported by the loop. Third, the agent re-tested augmentation strategies on a new dataset despite their having failed on the previous one (an act of non-greedy behaviour), but this required the human-declared scale transition to unlock.

\paragraph{Evidence of a residual inherent bias.}
Two observations suggest the bias is not purely workflow-induced. First, the risk-aversion pattern after a bold failure persisted for several hypotheses even when the workflow explicitly allowed another bold attempt; this is the agent choosing safe proposals, not the rule forcing it to. Second, the agent anchored repeatedly on a small set of familiar, well-cited papers when searching the literature and rarely engaged with more recent or less-cited work; we attribute this to the agent retrieving from an internalised prior rather than to an instruction to prefer familiar sources.

\begin{figure}[t]
\centering
\includegraphics[width=\linewidth]{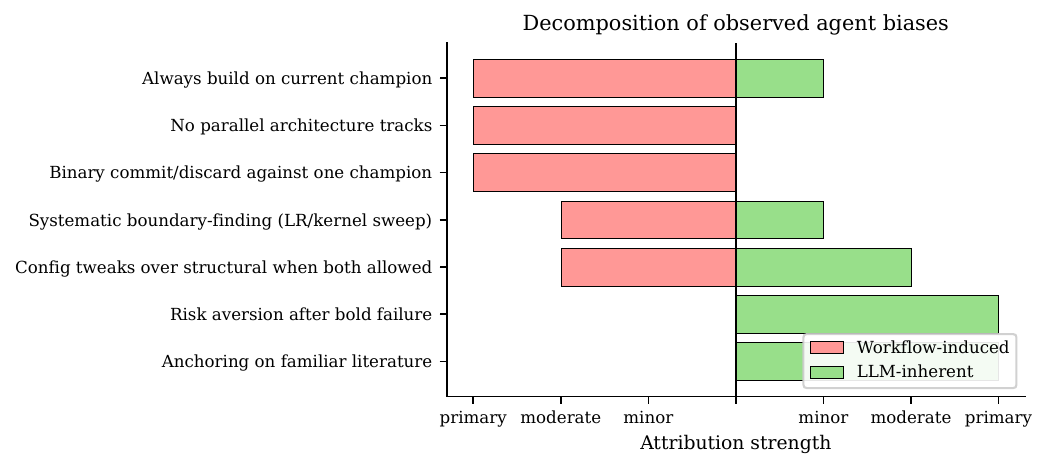}
\caption{Diverging-bar decomposition of the observed behavioural biases. Red bars to the left represent the component attributable to the workflow we imposed (which is structurally a greedy hill-climb); green bars to the right represent the residual component attributable to the agent's own tendencies. Every bias with a workflow-induced component is directly addressable by a workflow change; the residual component is smaller but harder to modify.}
\label{fig:bias_decomp}
\end{figure}

\paragraph{Decomposition.}
Table~\ref{tab:bias_decomposition} and Figure~\ref{fig:bias_decomp} summarise a qualitative decomposition of the observed behavioural biases into a workflow-induced component and a residual inherent component. The decomposition is not a controlled ablation; we report it as a structured hypothesis that a future multi-workflow study could test.

\begin{table}[h]
\centering
\caption{Qualitative decomposition of observed agent-behaviour biases into instruction-induced (workflow-imposed) and inherent-to-the-LLM components. Scale: $\bullet$ minor, $\bullet\bullet$ moderate, $\bullet\bullet\bullet$ primary. Positive behaviours (``retry in new context'', ``re-explore on scale transition'') are listed separately.}
\label{tab:bias_decomposition}
\small
\begin{tabular}{lcc}
\toprule
Behaviour & Instruction-induced & LLM-inherent \\
\midrule
Always build on current champion         & $\bullet\bullet\bullet$ & $\bullet$ \\
No parallel architecture tracks          & $\bullet\bullet\bullet$ & --- \\
Binary commit/discard against one champion & $\bullet\bullet\bullet$ & --- \\
Systematic boundary-finding (LR sweep, kernel sweep) & $\bullet\bullet$ & $\bullet$ \\
Config tweaks over structural when both allowed   & $\bullet\bullet$ & $\bullet\bullet$ \\
Risk aversion after bold failure         & ---                     & $\bullet\bullet\bullet$ \\
Anchoring on familiar literature         & ---                     & $\bullet\bullet\bullet$ \\
\midrule
\multicolumn{3}{l}{\emph{Positive behaviours (not biases)}} \\
Retry failed ideas in new context        & ---                     & $\bullet\bullet\bullet$ \\
Re-explore training recipe on scale transition & ---               & $\bullet\bullet\bullet$ \\
Ablation after success                   & $\bullet$               & $\bullet\bullet$ \\
\bottomrule
\end{tabular}
\end{table}

\paragraph{Implication.}
Our central behavioural finding is that the workflow designed around the agent is at least as consequential as the agent's capabilities for the research behaviour that appears over long horizons. A different workflow (one that maintains a small population of parallel architectures, budgets mandatory moonshot hypotheses, supports explicit backtracking, and enforces a periodic diversity audit) would likely produce meaningfully different and plausibly better research behaviour with the same underlying agent. Improving workflows is, in our view, a more tractable lever than attempting to change the agent's own risk profile.

\section{Lessons and Recommendations}
\label{sec:lessons}

\subsection{What Worked}
\label{sec:lessons_worked}

Four workflow choices had visibly positive effect across the programme. A single authoritative instruction document, rewritten whenever the agent's behaviour drifted, kept the agent on task more reliably than any change to model or prompt temperature. A fixed per-hypothesis record template, requiring the agent to commit motivation, literature grounding, change description, budget and outcome to a persistent log, created a searchable body of prior reasoning that the agent itself used in later proposals; removing the template, briefly attempted for token efficiency, measurably degraded decision quality. Per-hypothesis source-control history provided clean rollback and, incidentally, a compact audit trail of the programme. Integrating training-telemetry tooling gave the agent the ability to interpret results with reference to training dynamics rather than final metrics alone.

\subsection{What Did Not Work}
\label{sec:lessons_did_not_work}

Several choices produced negative lessons. Unsupervised agent autonomy was fragile: the agent would sometimes decide a turn was finished after merely submitting a job, mis-classify queued jobs as running, or attempt to run training on the shared login host. Each failure mode was fixable, but only after it appeared, and the fix was almost always an instruction edit rather than a model change. A permissive research brief that did not require literature grounding produced, in the first phase, a hypothesis distribution dominated by configuration tweaks rather than mechanisms; adding a literal ``cite a paper or declare the hypothesis an internal extrapolation'' rule shifted the distribution noticeably. Token-rate limits disrupted long sessions until we aggressively pruned what the agent was asked to re-read each turn and introduced a compact synthesis document summarising the state of the programme. The greedy workflow produced incremental behaviour even when structural leaps were available. And the agent's post-failure risk aversion was self-reinforcing: without external prompting, the loop would not have escaped the conservative regime entered after the first bold failure.

\subsection{Recommendations for Future Autonomous-Research Loops}
\label{sec:lessons_recommendations}

We offer five candidate interventions, each targeting a behavioural failure mode observed in our programme. Because they are grounded in a single run, we state them as hypotheses for future multi-agent, multi-problem studies rather than validated prescriptions.

\emph{Diversify the proposal strategy.} Replace pure greedy hill-climbing with a mixed policy: a majority of hypotheses that build on the current champion, combined with a minority drawn from an explicit ``diversification'' mode that picks a direction not recently tried, a return to an earlier branch point, or a composition of two individually refuted ideas (in the spirit of novelty search~\citep{lehman2011novelty}). A small multi-armed bandit over a short list of modes is sufficient and costs essentially nothing to implement.

\emph{Budget moonshot hypotheses.} Require that every $k$-th hypothesis introduce a mechanism not previously attempted, irrespective of the agent's estimated probability of success. This directly counters the inherent post-failure risk aversion observed in Section~\ref{sec:behavior_bias}.

\emph{Support explicit architectural forks.} Allow the agent to maintain a small number of parallel champions representing distinct design philosophies, alternate hypothesis allocation between them, and reconcile after a fixed number of iterations per track. This addresses the structural limitation of the single-champion workflow.

\emph{Track regime metadata on every accepted finding.} Annotate each finding with the regime (scale, dataset, epoch budget, augmentation family) in which it was validated. On any regime transition, automatically schedule re-validation of the top findings from the previous regime. This eliminates the silent carry-over of prior-scale decisions that cost us multiple hypotheses at each scale transition.

\emph{Compact synthesis documents.} The per-hypothesis log grows past the agent's effective attention window over long programmes. A compact, agent-authored synthesis document (periodically regenerated and read once per session in addition to the full log) materially improved the coherence of later proposals.

\subsection{Cost and Effort}
\label{sec:lessons_cost}

For reproducibility we record approximate costs. The programme spans several weeks of wall-clock, with approximately 90\% of the elapsed time being idle-on-queue or training. Total GPU-hours, total LLM tokens, and human-researcher hours are tabulated in Appendix~\ref{sec:appendix_impl}. We make no claim that this loop is cheaper than a human researcher; our claim is only that the \emph{research work} was executed by the agent, with the human role confined to the interventions recorded in Table~\ref{tab:interventions}, and that the resulting behavioural record is worth studying on its own.

\section{Limitations and Open Questions}
\label{sec:limitations}

\paragraph{Single agent, single problem.}
Our analysis is grounded in a single run of a single agent on a single architectural problem. Several quantitative findings (the concentration of gain in the first accepted hypothesis, the specific length of the saturation plateau, the specific list of rediscoveries) may be problem-specific. We expect the \emph{qualitative} phenomena (three-regime productivity structure, tool-access phase transition, greedy anchoring, post-failure risk aversion, cross-scale anti-transfer) to recur across problems and agents, but cannot verify this without parallel studies on different frontier models and different research problems.

\paragraph{No counterfactual human control.}
We do not have a matched human researcher running the same problem on the same compute with the same hypothesis budget. The qualitative comparisons we draw to typical human research behaviour are therefore indicative rather than controlled. A matched human control would be expensive but highly informative, and is the single experiment most likely to strengthen or overturn our claims.

\paragraph{Workflow-design confounds.}
The specific workflow we imposed (greedy commit-or-discard, single champion, one job at a time) directly shapes the behaviour we observe. Our decomposition of the observed bias into a workflow-induced component and a residual inherent component (Section~\ref{sec:behavior_bias}) is qualitative, not a controlled ablation. A cleaner design would run several loop variants over the same problem and measure how sensitive the behavioural outcomes are to each workflow choice independently of the agent.

\paragraph{Tool-access confound.}
The transition from the first to the second phase of the programme coincided with several simultaneous changes: literature access was added, code editing was authorised, and the research brief was amended to encourage novel mechanisms. Attributing the productivity increase to literature access specifically, rather than to the simultaneous instruction changes, is not clean. A controlled ablation would stage each change independently.

\paragraph{Evaluation-protocol divergence.}
The three phases of the programme use somewhat different evaluation protocols (augmentation family, epoch budget, dataset difficulty). Some of the cross-scale anti-findings in Section~\ref{sec:quantitative_cross_scale} may be partially driven by protocol rather than scale. We did not always cleanly separate these.

\paragraph{Single long run.}
We report one contiguous multi-week run (Table~\ref{tab:compute}) rather than many replicates. Run-to-run variance (on the same problem, same agent, even the same seed context) is unknown. Because LLM proposals are stochastic, a different random trajectory could easily produce a different champion.

\paragraph{Open questions we did not resolve.}
Does the three-regime productivity structure reported in Section~\ref{sec:quantitative_regimes} generalise across problems, or is it an artefact of an architectural design space that happened to have a clean six-hypothesis discovery regime? Are the agent's independent rediscoveries evidence that general-purpose LLMs can substitute for domain intuition, or evidence that they are re-emitting patterns from their training data? What fraction of the long second-phase plateau is genuine saturation of the design space versus behavioural risk aversion that a better-designed loop would override? And how does the loop scale with problem difficulty: does the qualitative structure persist at a scale where a single run costs orders of magnitude more? We leave all of these to future work.

\section{Conclusion}
\label{sec:conclusion}

We gave a single general-purpose large language model the four research affordances it minimally needed (source and experiment management, experiment tracking, literature access, and a persistent structured working memory) and let it run a long-horizon architectural research programme with only bounded, strategic human intervention. We analysed the resulting behavioural trace.

Three findings organise the analysis. First, long-horizon autonomous research has a visible phase structure that is invisible at short-horizon benchmark time scales: a brief regime of rapid, easy gains; a long saturation plateau in which the agent proposes plausible variations that almost never help; and a recovery regime that is only unlocked when the agent's action surface is expanded. In our setting the saturation plateau, and not raw reasoning capability, was the defining failure mode.

Second, the agent's observed preference for greedy, incremental hypotheses appeared primarily workflow-induced. The commit-or-discard evaluation rule we imposed is isomorphic to greedy hill-climbing, and a capable agent faithfully executes it. A residual component of the observed bias is attributable to the agent itself (risk aversion after bold failures, anchoring on well-cited familiar work, success-chasing), but in this run the larger lever was workflow design, not model capability.

Third, within its limits the agent is a useful collaborator. It proposes, implements, and cleanly ablates hypotheses; it writes dense, honest per-experiment records that a human can audit; and, when the workflow permits, it can surface and adapt literature-grounded mechanisms and, occasionally, resist a transferred intuition that does not hold in an unfamiliar regime.

From these observations we draw a practical suggestion for designers of future autonomous-research systems: invest in the workflow. Diversified proposal, budgeted moonshot hypotheses, explicit architectural forks, and regime-aware re-validation are each low-cost changes to the loop that directly target failure modes we observed, and we offer them as testable hypotheses rather than validated prescriptions. We invite the community to test these predictions on different problems, different agents, and different workflow designs. Autonomous agents are not substitutes for human researchers in the near term; the more interesting question is what workflow turns a flawed-but-tireless agent into a productive research collaborator. This paper is one data point in what we hope will be a broader empirical programme.

\bibliographystyle{plainnat}
\bibliography{references}

\clearpage
\appendix
\section{Implementation Details}
\label{sec:appendix_impl}

The main text described the loop at the level of research affordances rather than specific tools. This appendix records the stack used in the case study and the human-authored documents that bounded the agent's behaviour. We do not claim that any single component is essential; we list them so that other groups can either replicate or substitute. Anonymized supplementary materials (the research log, the per-hypothesis CSV used to generate every figure in this paper, and the figure-generation scripts) are provided alongside the submission; the training code and tracking-service exports will be released under a permanent DOI upon acceptance.

\subsection{Agent and Tooling}
\label{sec:appendix_impl_agent}

The case-study run used a single frontier instruction-tuned LLM exposed through its terminal-oriented command-line interface, so that all tool use is achieved through ordinary shell and file-system primitives (file read, file edit, shell command, search). Any comparable CLI-capable LLM should be a drop-in replacement. The specific model identifier and version tags at the time of each phase are recorded in the accompanying materials. No fine-tuning, RAG layer, or domain-specific scaffold was used; the agent's only persistent memory across sessions is the research log described below.

Tool access was managed through the Model Context Protocol (MCP). Three external services were exposed through MCP from the second phase onward: a preprint search and full-text service for arXiv, a documentation-question-answering service for GitHub repositories, and a model-and-dataset hub. Each is presented to the agent through a uniform tool interface so that literature retrieval, repository inspection, and dataset enumeration follow the same call pattern.

\subsection{Compute Infrastructure}
\label{sec:appendix_impl_compute}

Experiments ran on a shared HPC allocation managed by Slurm, with a mix of accelerator generations. Small-scale phases used a single GPU per job; the large-scale phase used two GPUs per job (capped by the per-user allocation limit at our site). The agent was explicitly prohibited from running training on interactive login hosts; this constraint was enforced by an instruction in the research brief and, after one observed violation in early Phase~1, by a wrapper script that rejected training-command invocations outside of a Slurm job context. Aggregate compute is reported in Table~\ref{tab:compute} in the main text.

A typical hypothesis cycle consisted of: literature query (a few seconds, optional), code or configuration edit (under a minute of agent wall-clock), Slurm submission (seconds), queue wait (highly variable, often the dominant component on shared infrastructure), training (minutes to days depending on phase), and result interpretation and log entry (seconds of agent time). The end-of-turn-after-submission discipline (described in Section~\ref{sec:system_principles}) ensured that the agent did not consume tokens while the job was queued or training.

\subsection{Source, Experiment, and Memory Management}
\label{sec:appendix_impl_state}

A single Git repository, with one branch per phase and one commit per accepted hypothesis. Rejected hypotheses were reverted from the working tree but remain reconstructible from the research log. Commit messages followed a fixed template (\texttt{phase: short title (hypothesis ID)}), giving a one-line audit trail directly inside source control.

Experiment tracking used a standard third-party experiment-tracking service, with one project per phase. Every Slurm job logged per-epoch training loss, validation top-1 and top-5, parameter count, FLOPs, and a fixed set of system statistics. The agent was expected to reference training curves (not only final metrics) when interpreting ambiguous results; we observed it doing so on several occasions where a final number was within noise but a divergent training curve disqualified the run.

Persistent memory took the form of a single Markdown research log containing one appended entry per hypothesis in the fixed template described in Section~\ref{sec:appendix_impl_template}. The agent was required to read this file in full at the start of every session (after an extended pause or context reset). On long programmes the log eventually exceeded the agent's effective attention window, at which point a compact agent-authored \emph{empirical-laws synthesis} document was introduced (regenerated periodically as a forced exercise in summarisation) and read once per session in addition to the log.

\subsection{Persistence and Control Hand-off}
\label{sec:appendix_impl_persist}

The agent process was kept alive across user disconnects in a persistent terminal session. To return control to the agent on job completion without requiring the agent to poll the scheduler (which would have consumed tokens and triggered the failure mode of mis-classifying queued jobs as running), a short shell utility watched for the appearance of the per-job log file's termination marker and re-prompted the agent in the same terminal session. We treat this as an implementation detail rather than a methodological contribution; any equivalent control-hand-off mechanism (a daemon, a cron entry, a job-completion hook) would serve the same purpose.

\subsection{Human-Authored Instruction Documents}
\label{sec:appendix_impl_instr}

Three human-authored documents bounded the agent's behaviour. Their exact wording (including revisions over the programme) is included in the supplementary materials. We here summarise their structure.

\paragraph{The research brief.} The brief is two-layered: a fuller research-programme document (\texttt{RESEARCH\_PROGRAM.md} in the supplementary) and a shorter agent-facing brief (\texttt{RESEARCH\_BRIEF.md}) that points to it. Together they specify: the scientific question (here, designing a channel-primary vision transformer under a pure channel-attention constraint); the design constraints (no spatial self-attention at any stage, parameter and FLOP budget per phase, dataset for each phase); the success criterion (validation top-1 on a fixed held-out set); and a list of \emph{illustrative} directions intended to seed the agent's literature exploration without prescribing the answer (see Section~\ref{sec:appendix_impl_directions} below). The brief was rewritten between phases to reflect transitions (Table~\ref{tab:interventions}); the rewritten version was always presented as the agent's authoritative single source of truth, and the previous version was archived.

\paragraph{The workflow template.} A separate document specifying the per-hypothesis loop (the algorithm in Algorithm~\ref{alg:loop}), the expected structure of each log entry, and explicit prohibitions: no scheduler polling, no training on the interactive login host, one experiment in flight at a time, no architectural changes without a corresponding source-control commit, and a budget check (parameter count and estimated FLOPs) before any submission. This document was updated only to add reliability constraints in response to observed autonomy failures (Section~\ref{sec:appendix_impl_failures}); the scientific content lived in the research brief.

\paragraph{The literature-grounding requirement.} A short rule, added at the Phase~1/1b boundary: every hypothesis must either cite a concrete prior result (a paper, repository, or model card) or explicitly declare itself an internal extrapolation. This rule is short but had a visible effect on the agent's behaviour: it shifted hypothesis selection from configuration tweaks toward code-level mechanism changes (Figure~\ref{fig:hypothesis_mix}). We attribute the effect to the rule forcing the agent to ground its proposal in a specific external reference rather than in pattern completion over its recent context.

\subsection{Per-Hypothesis Log Template}
\label{sec:appendix_impl_template}

Each entry in the research log follows a fixed seven-field template:

\begin{enumerate}[topsep=2pt,itemsep=2pt,leftmargin=*]
\item \textbf{Identifier and title.} A short symbolic identifier (\texttt{H1}, \texttt{P2-H7}, \texttt{H3-12}) and a one-line title naming the single variable changed.
\item \textbf{Motivation.} Two or three sentences stating the hypothesis to be tested and why it is worth testing now.
\item \textbf{Literature basis.} A specific citation (paper, repository, or model card) or an explicit declaration that the hypothesis is an internal extrapolation. The literature-grounding rule (Section~\ref{sec:appendix_impl_instr}) requires this field to be non-empty from Phase~1b onward.
\item \textbf{Change description.} The exact diff applied: which configuration values, which code paths. This field is the source of the \textsc{cfg}/\textsc{cod} change-type classification used throughout the paper.
\item \textbf{Budget.} Parameter count, FLOP estimate, and expected wall-clock; all three are recomputed by a budget-check utility before submission.
\item \textbf{Result.} Final validation metric, $\Delta$ vs.\ current champion, and a short interpretive sentence written by the agent after the run.
\item \textbf{Status.} One of \texttt{pushed}, \texttt{discarded}, \texttt{pending}, with the corresponding Git SHA if pushed.
\end{enumerate}

The full log is provided as a Markdown file in the supplementary materials. A tabular projection of all hypotheses appears in Appendix~\ref{sec:appendix_full_log}.

\subsection{Illustrative Directions in the Phase~3 Brief}
\label{sec:appendix_impl_directions}

The Phase~3 brief included a non-exhaustive list of five illustrative structural directions, intended to seed (not constrain) the agent's hypothesis generation: funnelling architectures with extra extreme-spatial-compression stages, factorised or low-rank channel attention, sparse top-$k$ channel attention, hierarchical local-plus-global channel attention, and linear or state-space channel attention. The brief explicitly framed this as a \emph{menu}, instructed the agent to synthesise novel ideas from the literature, and required a per-hypothesis self-audit of which novelty category the proposal fell into. We treat the menu as a documented direction-shaping intervention (Table~\ref{tab:interventions}), because the agent's Phase~3 hypothesis set ended up including adaptations of all five.

\subsection{Autonomy Failure Modes Encountered}
\label{sec:appendix_impl_failures}

Several classes of autonomy failure appeared during the programme and were each fixed once. We list them because they are likely to recur in any similar setup.

\begin{enumerate}[topsep=2pt,itemsep=2pt,leftmargin=*]
\item \textbf{Premature turn termination.} In early Phase~1 the agent would sometimes declare a turn finished after merely \emph{submitting} a Slurm job, without waiting for or interpreting the result. Fixed by an explicit ``end your turn after submission and wait for an external signal'' instruction (Section~\ref{sec:system_principles}) and the control-hand-off mechanism of Section~\ref{sec:appendix_impl_persist}.
\item \textbf{Scheduler mis-reading.} The agent sometimes classified queued jobs as running, leading to spurious result interpretations. Fixed by routing queue-status queries through a small wrapper that returned only \texttt{queued} / \texttt{running} / \texttt{finished}, eliminating the noisy free-text scheduler output.
\item \textbf{Login-host training.} Twice in early Phase~1 the agent attempted to start a training process on the interactive login host. Fixed by a wrapper script that rejected training-command invocations outside of a Slurm job context.
\item \textbf{Distributed-data-parallel unused-parameter crash.} The agent's H3-18 hypothesis (a Synthesizer-style operator that bypasses the key projection) crashed at iteration~0 because DDP detected an unused parameter. The agent diagnosed the unused projection itself, applied a one-line fix that preserves the parameter's gradient flow without changing the numerics, and resubmitted, without external intervention.
\item \textbf{Context-window saturation.} On the longest runs (multi-month elapsed time, hundreds of log entries) the research log eventually exceeded the agent's effective attention window. Fixed by introducing the compact agent-authored synthesis document described in Section~\ref{sec:appendix_impl_state}.
\item \textbf{Token-rate-limit interruptions.} Long sessions occasionally hit the LLM provider's token-rate limit. Fixed by aggressively pruning what the agent was asked to re-read each turn (notes on this are in the research brief), and by tolerating short stalls without retrying through the loop.
\end{enumerate}

Each failure is an instance of the same general pattern: a discipline that is implicit in a human researcher's habits (do not call a job ``done'' before its results are read, do not run training on a login node, summarise prior work rather than re-read it from scratch each turn) needed to be made explicit in the agent's instructions or enforced mechanically by the loop. We expect any future autonomous-research system to need analogous discipline-encoding work; the specific list above is intended as a starting checklist.

\subsection{Approximate Costs}
\label{sec:appendix_impl_costs}

Wall-clock: approximately ten weeks across the three phases (Table~\ref{tab:compute}), of which roughly 90\% was either queue wait or training and the remaining 10\% was agent activity. GPU-hours: of order $2.4$\,k aggregated across all phases, dominated by the large-scale ImageNet validation runs; the median substantive run was short ($\approx 2.5$\,h). LLM-token consumption: of order tens of millions of tokens across all phases (we did not log this systematically; the figure is a budget-based estimate). Human-researcher involvement: bounded to the interventions listed in Table~\ref{tab:interventions}, plus approximately one to two hours per day of passive oversight during active phases (monitoring for autonomy failures and unblocking when they occurred).

We make no claim that this loop is cheaper than a human researcher; our claim is that the \emph{research work} was executed by the agent, with the human role confined to the recorded interventions, and that the resulting behavioural record is worth studying on its own.

\subsection{Code and Data Availability}
\label{sec:appendix_impl_availability}

The supplementary materials accompanying this submission contain:

\begin{enumerate}[topsep=2pt,itemsep=2pt,leftmargin=*]
\item the two-layered research brief (\texttt{RESEARCH\_PROGRAM.md} and \texttt{RESEARCH\_BRIEF.md}), the workflow template (\texttt{AGENT\_INSTRUCTIONS.md}), and the tooling-configuration document (\texttt{TOOLING.md}), all in their final form;
\item the anonymized research log (Markdown), with one entry per hypothesis in the template of Section~\ref{sec:appendix_impl_template};
\item the agent-authored empirical-laws synthesis document referenced in Section~\ref{sec:appendix_impl_state}, and the agent-authored architecture-evolution summary referenced in Section~\ref{sec:case_study_output};
\item the per-hypothesis CSV used to generate every figure and the longtable in Appendix~\ref{sec:appendix_full_log};
\item the figure-generation scripts (Python, $<$~500 lines total);
\item the per-phase Slurm submission script templates and the control-hand-off watcher script (with site-specific identifiers stripped);
\item the architectural specification of the champion at the end of each phase, as YAML.
\end{enumerate}

The full training code (the model implementation, training driver, and per-run configurations of failed hypotheses) is omitted from the submission to preserve double-blind anonymity, and will be released under a permanent identifier upon acceptance. Re-running the architecture on the released training code reproduces the champion accuracy at each phase to within run-to-run variance; we did not budget for multi-seed replications of every accepted hypothesis, and this is one of the limitations discussed in Section~\ref{sec:limitations}.

\clearpage
\section{Per-Hypothesis Record}
\label{sec:appendix_full_log}

Table~\ref{tab:full_hyp_log} reproduces the full per-hypothesis record in a single tabular projection: one row per submitted hypothesis (and one row per dataset baseline), across all four phase markers (Phase 1, Phase 1b, Phase 2, Phase 3). The table is sorted chronologically within each phase. Identifiers prefixed \texttt{H} are CIFAR-10 hypotheses (Phase~1 and Phase~1b), identifiers prefixed \texttt{P2-H} are CIFAR-100 hypotheses (Phase~2), and identifiers prefixed \texttt{H3-} are ImageNet-1K hypotheses (Phase~3). The \emph{description} column is the agent's own one-line summary of the change, truncated to a fixed width for compactness; the full description, motivation, literature basis, change diff, budget, and outcome interpretation are in the supplementary research log (Section~\ref{sec:appendix_impl_availability}).

\begin{longtable}{@{}llp{0.46\textwidth}cccc@{}}
\caption{Full per-hypothesis record extracted from the research log. \emph{Description} is truncated to the first 58 characters of the agent's title; the full motivation, literature citation, change description, and outcome interpretation are in the supplementary log. \emph{Type} distinguishes hypotheses whose only substantive change was a configuration value (\textsc{cfg}) from those that introduced or modified code (\textsc{cod}); all Phase 3 hypotheses were structural and are labelled \textsc{cod}. \emph{Acc} is the validation top-1 (\%) at the dataset and protocol of that phase. \emph{$\Delta$} is the change vs.\ the running champion at the time the hypothesis was submitted, with the dataset baseline treated as the initial champion for each chain.}
\label{tab:full_hyp_log} \\
\toprule
ID & Ph. & Description & Type & Acc & $\Delta$ & Status \\
\midrule
\endfirsthead
\toprule
ID & Ph. & Description & Type & Acc & $\Delta$ & Status \\
\midrule
\endhead
\bottomrule
\multicolumn{7}{r}{\textit{continued on next page}} \\
\endfoot
\bottomrule
\endlastfoot
Baseline-C10 & 1 & Dataset baseline & -- & 69.67 & -- &  \\
H1 & 1 & Add CA MLP (CA\_MLP\_RATIO = [4.0, 4.0, 1.0] & cfg & 91.08 & +21.41 & pushed \\
H2 & 1 & Stage-2-only CA MLP (CA\_MLP\_RATIO = [0, 0, 1.0] & cfg & 89.97 & -1.11 & discarded \\
H3 & 1 & Multi-head channel attention (NUM\_HEADS = [1, 2, 4] & cfg & 89.95 & -1.13 & discarded \\
H4 & 1 & Deeper stage-2 backbone (DEPTH = [1, 2, 11] & cfg & 91.13 & +0.05 & pushed \\
H5 & 1 & Larger early-stage CA MLP (CA\_MLP\_RATIO = [8.0, 8.0, 1.\ldots & cfg & 91.71 & +0.58 & pushed \\
H6 & 1 & DW shortcut backbone (DW\_SHORTCUT\_BACKBONE = true & cfg & 92.52 & +0.81 & pushed \\
H7 & 1 & Remove stochastic depth (DROP\_PATH\_RATE = [0,0,0] & cfg & 92.98 & +0.46 & pushed \\
H8 & 1 & Shorter warmup (WARMUP\_EPOCHS = 5 & cfg & 93.00 & +0.02 & pushed \\
H9 & 1 & Higher learning rate (BASE\_LR = 1e-3 & cfg & 93.77 & +0.77 & pushed \\
H10 & 1 & Reduce weight decay (WEIGHT\_DECAY = 0.01 & cfg & 93.36 & -0.41 & discarded \\
H11 & 1 & Push LR further (BASE\_LR = 2e-3 & cfg & 93.96 & +0.19 & pushed \\
H12 & 1 & Push LR to 4e-3 (BASE\_LR = 4e-3 & cfg & 93.41 & -0.55 & discarded \\
H13 & 1 & 4-stage architecture (NUM\_STAGES=4, DIM\_EMBED=[64,128,2\ldots & cfg & 94.11 & +0.15 & pushed \\
H14 & 1 & 4-stage with stage-3 stride=1 (14×14 spatial, PATCH\_STRI\ldots & cfg & 93.98 & -0.13 & discarded \\
H15 & 1 & Reallocate depth: 1 stage-1 block → 1 stage-3 block (DEPT\ldots & cfg & 93.79 & -0.32 & discarded \\
H16 & 1 & Re-tune LR for 4-stage model (BASE\_LR = 3e-3 & cfg & 93.84 & -0.27 & discarded \\
H17 & 1 & Larger QKV kernel in stages 0-2 (KERNEL\_QKV=[5,5,5,3] & cfg & 93.39 & -0.72 & discarded \\
H18 & 1 & dw\_sep patch embedding at all 4 stages (PATCH\_EMBED\_ME\ldots & cfg & 93.47 & -0.64 & discarded \\
H19 & 1 & Deeper stage 0 (DEPTH=[2,2,3,7] & cfg & 93.55 & -0.56 & discarded \\
H20 & 1 & DEPTH=[1,2,3,8] + CA\_MLP\_RATIO=[8,8,3.3,1] (stage-3 dep\ldots & cfg & 93.68 & -0.43 & discarded \\
H21 & 1 & Disable Mixup+CutMix (AUG.MIXUP=0, AUG.CUTMIX=0 & cfg & 94.79 & +0.68 & pushed \\
H22 & 1 & Disable all DeiT augmentation (no RandAugment, no RandomE\ldots & cfg & 94.24 & -0.55 & discarded \\
H23 & 1 & Disable label smoothing (MODEL.LABEL\_SMOOTHING=0.0 & cfg & 94.92 & +0.13 & pushed \\
H24 & 1 & Reduce RandomErase probability (REPROB=0.1 & cfg & 94.71 & -0.21 & discarded \\
H25 & 1 & Mild drop path in deepest stage (DROP\_PATH\_RATE=[0,0,0,\ldots & cfg & 94.96 & +0.04 & pushed \\
H26 & 1 & Redistribute depth & cfg & 94.84 & -0.12 & discarded \\
H27 & 1 & Longer warmup (WARMUP\_EPOCHS=10 & cfg & 95.06 & +0.10 & pushed \\
H28 & 1 & Even longer warmup (WARMUP\_EPOCHS=15 & cfg & 94.99 & -0.07 & discarded \\
H29 & 1 & Progressive drop path [0, 0.01, 0.02, 0.05] & cfg & 95.09 & +0.03 & pushed \\
H30 & 1 & Layer Scale (CaiT-style, init=1e-4 & cod & 95.10 & +0.01 & pushed \\
H31 & 1 & Attention dropout in stage 3 (ATTN\_DROP\_RATE=[0,0,0,0.1] & cfg & 94.98 & -0.12 & discarded \\
H32 & 1 & DW shortcut placed between attention and MLP (DW\_SHORTCU\ldots & cfg & 95.09 & -0.01 & discarded \\
H33 & 1 & SiLU (Swish) activation instead of GELU & cfg & 94.91 & -0.19 & discarded \\
H34 & 1 & Higher weight decay (WEIGHT\_DECAY=0.1 & cfg & 94.99 & -0.11 & discarded \\
H35 & 1 & Lighter RandAugment (rand-m7 & cfg & 95.03 & -0.07 & discarded \\
H36 & 1 & More depth at stage 0 (DEPTH=[2,2,3,7] & cfg & 94.79 & -0.31 & discarded \\
H37 & 1 & Light Mixup (MIXUP=0.2 & cfg & 94.83 & -0.27 & discarded \\
H38 & 1 & All-dw\_sep patch embedding (['dw\_sep','dw\_sep','dw\_se\ldots & cfg & 94.80 & -0.30 & discarded \\
H39 & 1 & Deep pure CA & cfg & 94.03 & -1.07 & discarded \\
H40 & 1 & AdamW $\beta_2$ =0.98 & cfg & 94.74 & -0.36 & discarded \\
H41 & 1 & Layer Scale init=1e-5 & cfg & 94.98 & -0.12 & discarded \\
H42 & 1 & QK-Normalize (L2-normalize Q and K before dot product & cod & 94.22 & -0.88 & discarded \\
H43 & 1b & Learnable Channel Pair Bias (CPB & cod & 94.97 & -0.13 & discarded \\
H46 & 1b & Auxiliary deep supervision at stage 2 (intermediate class\ldots & cod & 95.70 & +0.60 & pushed \\
H47 & 1b & Per-stage learned attention temperature (log-parameterize\ldots & cod & 95.82 & +0.12 & pushed \\
H48 & 1b & Gated attention output (scalar sigmoid gate, channel mode & cod & 95.75 & -0.07 & discarded \\
H49 & 1b & Larger DW shortcuts (5×5 kernels, uniform across all stag\ldots & cfg & 95.71 & -0.11 & discarded \\
H50 & 1b & Dual auxiliary deep supervision at stages 1 and 2 & cod & 95.85 & +0.03 & pushed \\
H51 & 1b & Convolutional Position Encoding (CPE) at each stage & cod & 95.95 & +0.10 & pushed \\
H52 & 1b & Larger CPE kernel (5×5 & cod & 96.11 & +0.16 & pushed \\
H53 & 1b & CPE with 7×7 kernel & cod & 95.85 & -0.26 & discarded \\
H54 & 1b & SwiGLU MLP (replace GELU with SwiGLU gate in CA MLP blocks & cod & 96.23 & +0.12 & pushed \\
H55 & 1b & Per-stage CPE kernel sizes [3, 3, 5, 5] & cod & 96.24 & +0.01 & discarded \\
H56 & 1b & DW shortcut kernel verified [3,3,3,3] on H54 base & cfg & 96.48 & +0.25 & pushed \\
H57 & 1b & RMSNorm on H56 clean base & cod & 96.25 & -0.23 & discarded \\
H58 & 1b & Multi-head channel attention NUM\_HEADS [1,2,4,1] & cfg & 96.12 & -0.36 & discarded \\
H59 & 1b & Stronger auxiliary loss weights & cod & 96.42 & -0.06 & discarded \\
H60 & 1b & Wider stage 2 MLP (ratio 4→6), fewer blocks (3→2 & cfg & 96.38 & -0.10 & discarded \\
H62 & 1b & Depth redistribution [1,3,3,6] & cfg & 96.37 & -0.11 & discarded \\
H63 & 1b & LayerScale 1e-3 (10x larger initial scale & cod & 96.20 & -0.28 & discarded \\
H64 & 1b & Triple auxiliary supervision & cod & 96.59 & +0.11 & pushed \\
H65 & 1b & Stronger aux weights & cod & 96.48 & -0.11 & discarded \\
H67 & 1b & Intra-stage-3 mid-block auxiliary head (after block 4 of 7 & cod & 96.37 & -0.22 & discarded \\
Baseline-C100 & 2 & Dataset baseline & -- & 81.34 & -- &  \\
P2-H2 & 2 & Stage-3 depth-width tradeoff & cfg & 84.11 & +2.77 & discarded \\
P2-H3 & 2 & Group channel attention (G=4) at stage 3 + deeper network & cod & 81.31 & -0.03 & discarded \\
P2-H4 & 2 & Cross-stage channel feature fusion (DuoFormer-inspired & cod & 81.72 & +0.38 & pushed \\
P2-H5 & 2 & No stochastic depth at 100ep (DROP\_PATH=[0,0,0,0] & cfg & 81.12 & -0.60 & discarded \\
P2-H6 & 2 & Label smoothing LS=0.1 at 100ep & cfg & 82.02 & +0.30 & pushed \\
P2-H7 & 2 & Mild Mixup=0.4 at 100ep & cfg & 82.79 & +0.77 & pushed \\
P2-H8 & 2 & Differential channel attention & cod & 81.88 & -0.91 & discarded \\
P2-H9 & 2 & Intra-stage-3 mid-block auxiliary supervision & cod & 82.80 & +0.01 & pushed \\
P2-H10 & 2 & Periodic CPE re-injection within stage 3 (interval=4 & cod & 82.88 & +0.08 & pushed \\
P2-H11 & 2 & Larger Q/K/V projection kernel at stage 3 (3×3 → 5×5 & cfg & 83.16 & +0.28 & pushed \\
P2-H12 & 2 & Full-field Q/K/V projection kernel at stage 3 (5×5 → 7×7 & cfg & 83.11 & -0.05 & discarded \\
P2-H13 & 2 & 5×5 Q/K/V projection kernel at stage 2 as well & cfg & 82.82 & -0.34 & discarded \\
P2-H14 & 2 & 5×5 DW shortcut kernel at stage 3 & cfg & 83.17 & +0.01 & pushed \\
P2-H15 & 2 & Wider stage-3 embedding (576→640) with fewer depth (12→10 & cfg & 83.18 & +0.01 & pushed \\
P2-H16 & 2 & Cross-stage spatial input fusion (stage-2 features → stag\ldots & cod & 82.72 & -0.46 & discarded \\
P2-H17 & 2 & Grouped channel attention at stage 3 (G=4 groups of 160 c\ldots & cod & 82.98 & -0.20 & discarded \\
P2-H18 & 2 & Sigmoid channel attention at stage 3 & cod & 82.97 & -0.21 & discarded \\
P2-H19 & 2 & Value residual for stage-3 channel attention & cod & 83.11 & -0.07 & discarded \\
P2-H20 & 2 & Dual-scale channel attention (GAP branch & cod & 83.16 & -0.02 & discarded \\
P2-H21 & 2 & Learned channel-pair attention prior (CHANNEL\_PAIR\_BIAS & cod & 82.89 & -0.29 & discarded \\
P2-H22 & 2 & All-stage cross-stage fusion (add stage 0 & cod & 83.30 & +0.12 & pushed \\
P2-H23 & 2 & Second-order statistics pooling in cross-stage fusion & cod & 83.30 & +0.00 & discarded \\
P2-H24 & 2 & Attentive channel gating (output-side gate on stage-3 cha\ldots & cod & 83.13 & -0.17 & discarded \\
P2-H25 & 2 & Decoupled V projection & cod & 82.82 & -0.48 & discarded \\
P2-H26 & 2 & Per-stage periodic CPE re-injection & cod & 83.37 & +0.07 & pushed \\
P2-H27 & 2 & Stage-3 CPE interval 4→3 (denser spatial refresh in deepe\ldots & cod & 82.74 & -0.63 & discarded \\
P2-H28 & 2 & 5×5 Q/K at stage-2 (extend stage-3 winning kernel size to\ldots & cfg & 82.80 & -0.57 & discarded \\
P2-H29 & 2 & Register tokens for stage-3 channel attention (K=4 & cod & 82.70 & -0.67 & discarded \\
P2-H30 & 2 & Stage-1 periodic CPE re-injection (interval=1 & cod & 82.51 & -0.86 & discarded \\
P2-H31 & 2 & 7×7 CPE kernel & cod & 83.03 & -0.34 & discarded \\
P2-H32 & 2 & Depth redistribution DEPTH=[1,2,5,10]→[1,2,6,9] & cfg & 83.17 & -0.20 & discarded \\
Baseline-IN1K & 3 & Dataset baseline & -- & 77.65 & -- &  \\
H3-1 & 3 & Sparse top-k channel attention at stage-3 (k=64 of C=640 & cod & 77.48 & -0.17 & discarded \\
H3-2 & 3 & Dual-resolution channel attention at stage-3 (G=32 super-\ldots & cod & 77.50 & -0.15 & discarded \\
H3-3 & 3 & Cross-stage channel attention bridge (stage-2 → stage-3 & cod & 77.55 & -0.10 & discarded \\
H3-4 & 3 & Agent attention at stage-3 (factorised low-rank CA via M=\ldots & cod & 77.38 & -0.27 & discarded \\
H3-5 & 3 & Hierarchical local + global CA at stage-2 (within-group f\ldots & cod & 77.45 & -0.20 & discarded \\
H3-6 & 3 & Funneling tail-block at 3×3 spatial pool (purely additive\ldots & cod & 77.59 & -0.06 & discarded \\
H3-8 & 3 & Linear-attention funneling tail at full stage-3 spatial (\ldots & cod & 62.93 & -14.72 & discarded \\
H3-9 & 3 & Linformer-style projected channel attention funnel + shar\ldots & cod & 77.79 & +0.14 & pushed \\
H3-10 & 3 & Stacked Linformer-style projected channel attention funne\ldots & cod & 77.75 & -0.04 & discarded \\
H3-11 & 3 & Parallel hybrid funnel & cod & 77.58 & -0.21 & discarded \\
H3-12 & 3 & Linformer applied to ALL 10 stage-3 in-trunk CA blocks (L\ldots & cod & 77.91 & +0.12 & pushed \\
H3-13 & 3 & Push Linformer wider & cod & 77.65 & -0.26 & discarded \\
H3-14 & 3 & Per-block Linformer rank schedule within stage-3 & cod & 77.67 & -0.24 & discarded \\
H3-15 & 3 & Sparse top-k WITHIN the Linformer rank-128 supertoken bas\ldots & cod & 77.54 & -0.37 & discarded \\
H3-16 & 3 & Gated parallel FULL-RANK CA pathway alongside Linformer a\ldots & cod & 77.78 & -0.13 & discarded \\
H3-17 & 3 & Sparsemax replacement for softmax inside stage-3 Linforme\ldots & cod & 77.74 & -0.17 & discarded \\
H3-18 & 3 & Synthesizer-Dense channel attention at stage-3 (REPLACES \ldots & cod & 77.39 & -0.52 & discarded \\
\end{longtable}

\end{document}